%% file: main.tex
\documentclass{article}

\PassOptionsToPackage{sort&compress,numbers}{natbib}
\usepackage[preprint]{neurips_2021}

\usepackage[utf8]{inputenc} %
\usepackage[T1]{fontenc}    %
\usepackage{url}            %
\usepackage{nicefrac}       %
\usepackage{microtype}      %
\usepackage{graphicx}
\usepackage{tikz}

\usepackage[pagebackref=true,breaklinks=true,colorlinks,bookmarks=false]{hyperref}

\usepackage{amsfonts}
\usepackage{amsmath}
\usepackage{amssymb}

\usepackage[noabbrev,capitalize]{cleveref}

\usepackage{booktabs}
\usepackage{multirow}
\usepackage{support-caption} %
\usepackage{subcaption}

\input{custom_macro}

\usepackage[accsupp]{axessibility}  %

\begin{document}

\title{ECCV Caption: Correcting False Negatives by Collecting Machine-and-Human-verified Image-Caption Associations for MS-COCO}

\author{Sanghyuk Chun, Wonjae Kim, Song Park, Minsuk Chang$^\diamondsuit$, Seong Joon Oh$^\clubsuit$\\
\\
NAVER AI Lab\\
$\diamondsuit$ Now at Google Research \quad $\clubsuit$ Now at University of Tübingen
}

\maketitle
\begin{abstract}
Image-Text matching (ITM) is a common task for evaluating the quality of Vision and Language (VL) models. However, existing ITM benchmarks have a significant limitation. They have many missing correspondences, originating from the data construction process itself. For example, a caption is only matched with one image although the caption can be matched with other similar images and vice versa. To correct the massive false negatives, we construct the Extended COCO Validation (ECCV) Caption dataset by supplying the missing associations with machine and human annotators. We employ five state-of-the-art ITM models with diverse properties for our annotation process. Our dataset provides $\times$3.6 positive image-to-caption associations and $\times$8.5 caption-to-image associations compared to the original MS-COCO. We also propose to use an informative ranking-based metric mAP@R, rather than the popular Recall@K (R@K). We re-evaluate the existing 25 VL models on existing and proposed benchmarks. Our findings are that the existing benchmarks, such as COCO 1K R@K, COCO 5K R@K, CxC R@1 are highly correlated with each other, while the rankings change when we shift to the ECCV mAP@R. Lastly, we delve into the effect of the bias introduced by the choice of machine annotator. Source code and dataset are available at \texttt{\url{https://github.com/naver-ai/eccv-caption}}
\end{abstract}

\input{1_introduction}
\input{2_relworks}
\input{3_dataset}
\input{4_benchmark}
\input{5_discussion}
\input{6_conclusion}
\input{7_acknowledgement}
\section*{Appendix}
\input{99_appendix}

\bibliography{reference}
\bibliographystyle{unsrtnat}
\end{document}

%% file: custom_macro.tex
\usepackage{xcolor}
\usepackage{xspace}
\def\onedot{.\xspace}
\def\eg{\emph{e.g}\onedot} 
\def\ie{\emph{i.e}\onedot}

\def\etal{\emph{et al}\onedot}

\usepackage{pifont}
\definecolor{darkergreen}{RGB}{21, 152, 56}
\definecolor{red2}{RGB}{252, 54, 65}

\newcommand{\hc}[1]{\textbf{\textcolor{red}{#1\xspace}}}

\newcommand{\ourdatasetfull}{Extended COCO Validation (ECCV) Caption dataset\xspace}
\newcommand{\ourdataset}{ECCV Caption\xspace}
\newcommand{\eccv}{ECCV\xspace}

\newcommand{\recallatk}{Recall@$k$\xspace}
\newcommand{\rk}{R@$k$\xspace}
\newcommand{\recallatone}{Recall@1\xspace}
\newcommand{\rone}{R@1\xspace}
\newcommand{\rfive}{R@5\xspace}
\newcommand{\rten}{R@10\xspace}
\newcommand{\map}{mAP\xspace}
\newcommand{\mapr}{mAP@$R$\xspace}

%% file: 1_introduction.tex
\section{Introduction}
\begin{figure}[t]
    \centering
    \includegraphics[width=\linewidth]{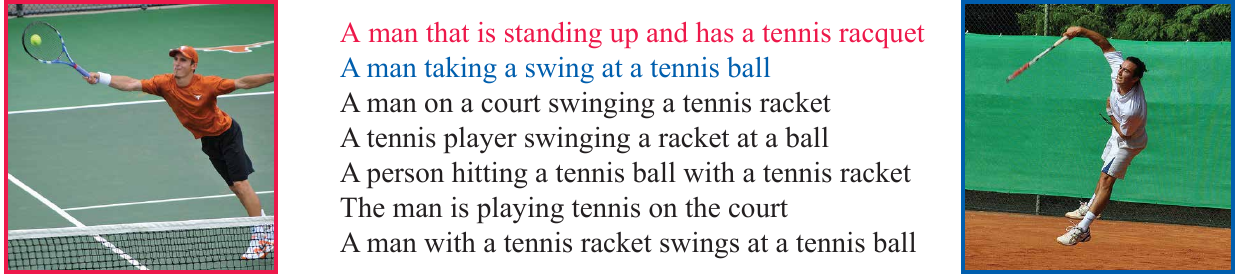}%
    \caption{\small {\bf Inherent multiplicity of correspondences in MS-COCO Caption.} While any image-caption pair above makes sense (positive pair), only \textcolor{red}{red} and \textcolor{blue}{blue} image-caption pairs are marked as positive in MS-COCO Caption.}%
    \label{fig:multiplicity}%
\end{figure}

Image-caption aligned datasets (\eg, MS-COCO Caption \cite{lin2014microsoft, chen2015microsoft}, Flickr30k \cite{plummer2015flickr30k}, Conceptual Caption \cite{sharma2018conceptual, changpinyo2021conceptual}) have become \textit{de-facto} standard datasets for training and evaluating Vision-Language (VL) models. Particularly, Image-to-Text Matching (ITM) tasks 
\cite{frome2013devise, young2014image, kiros2014unifying, faghri2018vsepp, gu2018look, lee2018scan, huang2018learning, Li2019VSRN, song2019pvse, wehrmann2019language, wu2019unified, Wang2020CVSE, chen2020adaptive, Diao2021SGRAF, chun2021pcme, chen2021vseinfty, huang2021learning, biten2022image}
are widely used benchmarks for evaluating a VL model. 
The existing ITM benchmark datasets
are built by annotating captions (by alt-texts \cite{sharma2018conceptual, changpinyo2021conceptual, clip}, web crawling \cite{desai2021redcaps}, or human annotators \cite{chen2015microsoft}) for each image without considering possible associations with other images in the dataset. The collected image-caption pairs are treated as the only positives in the dataset, while other pairs are considered the negatives. However, in practice, there exists more than one caption to describe one image. For example, the description ``A man that is standing up and has a tennis racquet'' may describe multiple images with tennis players equally well (\cref{fig:multiplicity}). We have observed that the number of missing positives is tremendous; there exist $\times$3.6 positive image-to-caption correspondences and $\times$8.5 caption-to-image correspondences than the original MS-COCO dataset.

While the huge number of false negatives (FNs) in VL datasets is potentially sub-optimal for training VL models, it is downright detrimental for evaluation. For example, the small number of positive correspondences of image-caption-aligned datasets limits the evaluation metrics.\footnote{In MS-COCO Caption, a caption is only matched to one image, and an image is matched to five captions. Other datasets usually have one caption for each image.}
In other tasks, such as image retrieval \cite{cub, cars, sop, inshop}, 
the positives and negatives are defined by class labels; hence, the number of possible matched items is large enough to measure precision or mean average precision (\map) metrics. On the other hand, because existing ITM benchmarks only have one positive correspondence for each item, they are only able to use recall-based metrics (\eg, \recallatk) that are known to be less informative than the precision- or ranking-based evaluation metrics \cite{musgrave2020metric}. In this paper, we focus on correcting the FNs in the evaluation dataset and the recall-based evaluation metrics to make a fair comparison of VL models.

As our first contribution, we correct the FNs in MS-COCO Caption by constructing \ourdatasetfull. We annotate whether each MS-COCO image-caption pair is positive with human workers. The labor cost for this process scales quadratically with the size of the dataset (\eg, MS-COCO has 76B possible image-caption pairs, while the number of images is only 123K). Since verifying every possible image-text pair is not scalable, we subsample the queries in the dataset and reduce the number of candidates for positive matches with the machine-in-the-loop (MITL) annotation process. MITL lets a model reduce the number of candidate positives; then human annotators evaluate the machine-selected candidates. We employ five state-of-the-art ITM models with distinct properties as machine annotators; CLIP \cite{clip}, ViLT \cite{kim2021vilt}, VSRN \cite{Li2019VSRN}, PVSE \cite{song2019pvse}, and PCME \cite{chun2021pcme}.
After post-processing, \ourdataset contains 1,261 image queries (originally 5,000) but with 17.9 positive captions per image query on average (originally 5). It also contains 1,332 caption queries (originally 25,000) with 8.5 positive images per caption (originally 1).

While the use of a machine annotator is inevitable for the sake of scalability, the choice of a particular model may bias the dataset towards the specifics of the model. This can be problematic because different models show different filtered results to the human annotators, which brings the impartialness of the annotated dataset towards any particular model to the surface. In other words, the MITL annotations are not stable across model choices. Our studies show that the underlying ML model conditions the annotated dataset towards favoring certain models over the others. Therefore, this practice could lead to the danger of biased evaluation results using such datasets. We show that the rankings among the VL models can be arbitrarily shifted by modifying the underlying ML model. Our study also shows that using multiple machine annotators can alleviate machine bias in dataset construction. We note that the findings are applicable to a wide range of tasks in which users put labels on samples from a long list of candidate classes; our task is a special case of such a framework.

A similar MITL approach 
for expanding the positive matches was also employed by Parekh \etal \cite{parekh2020crisscrossed}, resulting in the dataset CrissCrossed Caption (CxC). However, CxC focuses on scoring the text-to-text similarities, resulting in many missing positives in the text-to-image relationship. Furthermore, CxC only employs one language-based machine annotator, which can lead to a biased dataset as our observation. Our \ourdataset focuses on the inter-modality relationship and utilizes five ITM methods to avoid biased dataset construction.
As another attempt to correct COCO benchmark, Chun \etal \cite{chun2021pcme} annotate pseudo-positives by using the COCO instance classes, called Plausible Match (PM). For example, both images in \cref{fig:multiplicity} contain the same object class, ``tennis racket''. Hence, the red and blue captions are considered positives for both red and blue images. Although PM items can detect most of the false negatives, it also introduces many false positives.
Compared to PM \cite{chun2021pcme} which relies on noisy proxies for correspondence, we correct the missing false negatives with ``human ground truths'' with the help of machine annotations. All in all, our dataset results in a higher recall than CxC and high precision than PM.

We not only fix FNs but also evaluation metrics. We argue that \rone can overestimate the model performance by focusing only on the accuracy of the top-1 item rather than the rest of the items. Instead, we propose to use better ranking-based evaluation metrics, \mapr \cite{musgrave2020metric}. Our human study shows that \mapr is more aligned to humans than \recallatk.
Now that the FNs are corrected in the evaluation sets and the evaluation metric is fixed, we re-examine the known ranking of 25 state-of-the-art VL models evaluated in the COCO Caption.
We have observed that COCO 5K \rone\&\,\rfive, and CxC \rone are highly correlated (larger than 0.87 Kendall's rank correlation $\tau$).
On the other hand, we observe that the rankings across methods measured by \mapr on \ourdataset and COCO 1K \rone are less well-correlated ($\tau$=0.47). 
This confirms the observation by Musgrave \etal \cite{musgrave2020metric} and Chun \etal \cite{chun2021pcme} on class-based datasets.

Our contributions are as follows.
(1) We discover the false negative (FN) problem and quantify the exact number of wrong labels in MS-COCO. There exist $\times$3.6 positive image-to-caption associations and $\times$8.5 caption-to-image associations compared to the original MS-COCO.
(2) We construct a corrected ITM test dataset, \textbf{\ourdataset}, to avoid a wrong evaluation by FNs. We employ the machine-in-the-loop (MITL) annotation process to reduce the amount of human verification, resulting in saving 99.9\% cost compared to the full exhaustive verification.
\ourdataset shares the same images and captions as the original MS-COCO; therefore, the existing methods can be evaluated on our dataset without additional training. We fix not only the annotations but also the evaluation metric. We propose to use \mapr, a more human-aligned metric than \rone for comparing model performances as shown in our human study.
(3) We re-evaluate 25 state-of-the-art VL models on our \ourdataset dataset based on \mapr instead of \recallatk.
In \cref{tab:main_results} and \cref{fig:bump_main}, we can observe that focusing on MS-COCO \rone will mislead the true ranking between the models (MS-COCO \rone and ECCV \mapr show a low correlation). Our observation aligns with Musgrave \etal \cite{musgrave2020metric} and Chun \etal \cite{chun2021pcme}; focusing on \rone can mislead the true rankings between models.
(4) We provide a detailed analysis of the constructed dataset and the model bias. In particular, we focus on avoiding potential model biases in the proposed dataset by employing multiple models. Our analysis shows that our design choice is effective in solving the model bias.

%% file: 2_relworks.tex
\section{Related Works}

\subsection{Noisy many-to-many correspondences of image-caption datasets}

There have been a few attempts to introduce many-to-many or noisy correspondences for VL datasets. Parekh \etal \cite{parekh2020crisscrossed} construct a CrissCrossed Caption (CxC) dataset by employing a similar MITL approach to ours. However, CxC focuses on intra-modality similarity, particularly text-to-text. They employed the Universal Sentence Encoder \cite{cer2018universal} and average bag-of-words (BoW) based on GloVe embeddings \cite{pennington2014glove}, while 
we directly focus on the inter-modality relationships and utilizes 
powerful ITM methods \cite{clip, kim2021vilt, Li2019VSRN, song2019pvse, chun2021pcme}
to select candidates for validation by humans. CxC contains human ratings for 89,555 image-to-caption associations, among which 35,585 are positive, $\times$1.4 more positive relationships than 25,000 in COCO Caption. We show that the additional positives by CxC are precise, but their annotations still have many missing positives (\ie, high precision but low recall), resulting that \rone on CxC perfectly preserves the rankings of VL models on COCO 5K \rone. On the other hand, our \ourdataset has $\times$4.4 positives ($\times$3.6 image-to-caption correspondences and $\times$8.5 caption-to-image correspondences) compared to COCO Captions and roughly three times more positives compared to CxC. Furthermore, it is possible to measure \map on our dataset due to the abundance of positive pairs, unlike for CxC.

Another attempt by Chun \etal \cite{chun2021pcme} focused on precision rather than \rone by annotating the pseudo-positives in a fully algorithmic approach. The authors defined ``plausible matching (PM)'' items that have the same instance classes with the query image (or the image corresponding to the query caption) to annotate pseudo-positives. For example, both images in \cref{fig:multiplicity} contain the same instance class, ``tennis racket'', leading to the conclusion that the red and blue captions are marked as positives for both red and blue images. More precisely, two instances are PM if $y_1, y_2 \in \{0, 1\}^d$ differ at most $\zeta$ positions, where $d$ is the number of instance classes (\eg, for COCO, $d=80$). Using the class-based pseudo-positives, Chun \etal propose Plausible-Match R-Precision (PMRP) metric, an R-Precision \cite{musgrave2020metric} metric based on the PM policy. The authors propose to use multiple $\zeta$ (\eg, $\zeta \in \{0, 1, 2\}$) and report the average precision value. PM items can detect many missing false positives in the dataset, but we observe that most PM pseudo-positives are not actual positives (\ie, high recall but low precision) --- See \cref{tab:cxc_pm_pr}. We also observe that PMRP shows a low correlation to other evaluation metrics; PMRP is a noisy metric compared to others.

\subsection{Machine-in-the-loop (MITL) annotation}

Humans and machines complement each other in the annotation process as they have different comparative advantages. Humans are the ultimate source of true labels, but they are slow and prone to errors and biases \cite{snow-etal-2008-cheap,SorokinForsyth2008,ipeirotis2010quality}. Machines are highly scalable, but their generalizability to unseen samples is limited. Machines are also prone to their own versions of errors and biases \cite{MehrabiMorstatterSaxenaLermanGalstyan2021,scimeca2022wcst-ml}. MITL annotations have been designed to take the best of both worlds \cite{boykov2001interactive,settles2009active,xu2016deep,OpenImagesV5}. 

Depending on the required trade-off between annotation quality and efficiency, one may opt for either single-turn or multi-turn annotation pipeline. The latter serves for the maximal demand for annotation quality: humans and machines alternate to correct and learn from each other's annotations \cite{settles2009active,OpenImagesV5}. 
This is a widely used technique, the applications ranging from building a dictionary of cooking vocabularies~\cite{chang2018recipescape}, to supporting real-time screen-reading for blind people~\cite{Guo2016VizLensAR} and characterizing system failures~\cite{Nushi2018TowardsAA}.
Here, we focus on \textit{single-turn MITL annotations} to focus on the atomic building block for MITL pipelines in general. There are two types of the single-turn paradigm: machine-verified human annotations \cite{wu2006smartlabel,verma2017image} or human-verified machine annotations.
We focus on the latter, which are highly relevant for dealing with huge sources of data.

Under the human-verification framework, machines make label proposals for each image, focusing more on recall than precision \cite{andriluka2018fluid,openimagesv4}. 
Previous crowdsourcing research in human-computer interaction (HCI) had mainly focused on the annotation interface and its effects on the annotation~\cite{Kaplan2018StrivingTE, Song2018TwoTA, Chung2019EfficientEA}, or building a crowdsourcing workflow that leverages microtask pipelines~\cite{bernstein2010soylent, Kim2014CrowdsourcingSI}. We investigate the side effects of the model choice in the MITL annotation paradigm 
where machines provide candidate label proposals.

%% file: 3_dataset.tex
\section{\ourdataset Dataset Construction}
\label{sec:dataset_construction}

In this section, we describe 
\ourdataset construction details.
We annotate image-caption pairs in MS-COCO to solve the multiplicity of MS-COCO. However, the number of candidates is too huge for an exhaustive verification by humans: 76B for the whole dataset and 125M for the test split only. To reduce the amount of judgment by humans, we employ a single-turn machine-in-the-loop (MITL) annotation pipeline, containing three stages: (1) Filtering by machine annotators. (2) Judging the filtered relationships by MTurkers and additional verification by internal workers. (3) Post-processing and merging with CxC.

\subsection{Model candidates for machine annotators}
\label{subsec:model_candidates}

We choose five VL models with diverse properties to cover both diversity and practical relevance. 
The models use different text backbones (Bi-GRU \cite{cho2014properties}, Transformer \cite{vaswani2017attention}), visual backbones (ResNet-152 \cite{resnet}, Faster R-CNN \cite{ren2015faster}, ViT \cite{vit}), training objective functions, and training datasets as shown in \cref{tab:model_overview}.
We use the officially released pre-trained weights by the authors.
Specifically, we use the CutMix \cite{cutmix} pre-trained version for PCME
to match the retrieval performances with others, and CLIP ViT-B/32, the largest model at the time of our data construction. We describe more details of each method in 
\cref{subsec:machine_annotators_detail}.

\input{tables/models_overview}

We quantify the diversity of the models by measuring the differences in their retrieved items. We first retrieve the top 25 images for each model on the captions of the COCO Caption test split. We measure the similarities of the models in two different metrics. First, for every pair of models, we measure the Kendall rank correlation \cite{kendall1938new} between the two rankings of the retrieved items by the models. We observe that the models usually have low similarity ($\tau < 0.3$), except for PVSE and PCME. We additionally measure, for each pair of model $i$ and $j$, the average ranking of model $i$'s top-1 ranked item by model $j$. The top-1 items retrieved by the models are usually not included in the top-3 items by the others. These analyses show that the chosen models are diverse and the retrieved items do not correlate that much. The full results are shown in 
\cref{subsec:machine_annotators_kendall}.

\subsection{Crowdsourcing on Amazon Mechanical Turk}
\label{subsec:crowdsource-amt}

We crowdsource image-caption matches on Amazon Mechanical Turk (MTurk) platform. For the sake of scalability, we subsample 1,333 caption queries and 1,261 image queries from the COCO Caption test split. Since the number of all possible matches is still prohibitive 
(40M), 
we employ the filtering strategy to reduce the number of candidates for human verification. We pre-select top-5 captions and images retrieved by the five models. After we remove the duplicate pairs from the (1,261 $+$ 1,333) $\times$ 5 $\times$ 5 = 64,850 pairs, 46,424 pairs remain.

We package the task for human annotators into a series of Human Intelligence Tasks (HITs). Each HIT contains 20 pairs consisting of 18 machine-retrieved pairs to be annotated, 1 true positive (\ie, an \textit{original positive pair}), and 1 true negative (random pair, not in the top-25 of any model). The golden examples are used for the qualification process; if a submitted HIT contains wrong answers to the golden examples, we manually verify the HIT. For each image-caption pair candidate, workers can choose an answer among the choices ``100\% YES'', ``Partially YES, but'', ``Mostly NO, because'', and ``100\% NO''. We use four choices instead three-level (``YES'', ``Not Sure'', and ``NO'') to discourage workers from selecting ``Not Sure'' for all the questions. We have assigned 2,160 HITs, consisting of 43,200 pairs to be verified, to 970 MTurk workers.
The crowdsourcing details, including an example HIT, compensation details, worker statistics, and detailed statistics for each machine annotator are in 
\cref{subsec:mturk_results}

\subsection{Postprocessing MTurk annotations}
\label{subsec:postprocessing_annotations}

We observe that 21,995 associations among 43,200 associations are annotated as positives (``Yes'' or ``Weak Yes'').
We then filter out 18 meaningless captions (\eg, ``I am unable to see an image above''), 14 wrong captions found by workers (\eg, ``A group of birds flying above the beach'' for the image with many kites), and 1 duplicate image found in the training set.
The full list is in 
\cref{subsec:invalid_items}.

\input{tables/dataset_comparisons}

Using the 21,995 human-verified positives, we report precision and recall of the existing benchmarks. Let $t_i$ be the set of human-annotated positives for the query $i$ in \cref{subsec:crowdsource-amt} and $r_i$ be the set of positives for $i$ in the target dataset. Note that our human-annotated positives are based on the top-5 retrieval items of our machine annotators; if the original ``GT'' item is ranked in top-5 by none of the models, then $t_i$ will not include the original ``GT'' item. To prevent this, we use $r^\prime_i = r_i \cap h_i$ where $h_i$ is the set of human-verified pairs for the query $i$ (\ie, the top-5 items of our machine annotators). We filter out the case when $r^\prime_i = 0$ for preventing an ill-defined result\footnote{This filtering process is more critical to ``I2T'' precision results because the number of the original ``GT'' item per query is 1. We revise these result from the previous revision (\texttt{v3}).}.
We define precision and recall of a dataset as $Prec = \frac{1}{N}\sum_{i=1}^N \frac{| r^\prime_i \cap t_i |}{|r^\prime_i|}$ and $Recall =  \frac{1}{N}\sum_{i=1}^N 1 - \frac{|t_i ~\char`\\~ r^\prime_i|}{|t_i|}$.
\cref{tab:cxc_pm_pr} shows precision and recall of 
COCO Caption, CxC \cite{parekh2020crisscrossed}, and Plausible Match (PM) pseudo-positives \cite{chun2021pcme}. While COCO and CxC show high precisions, we observe that their recall is significantly low, around or less than 20\%. Evaluating models on such a low-recall dataset with the \rone metric can be highly misleading. A model may be able to retrieve good enough positive items which are not captured in the dataset, resulting in erroneously low \rone scores. On the other hand, around 60\% of the positives can be captured by PM, but around 60\% of pseudo-positives are correct.

\begin{figure}[t]
    \centering
    \includegraphics[width=\linewidth]{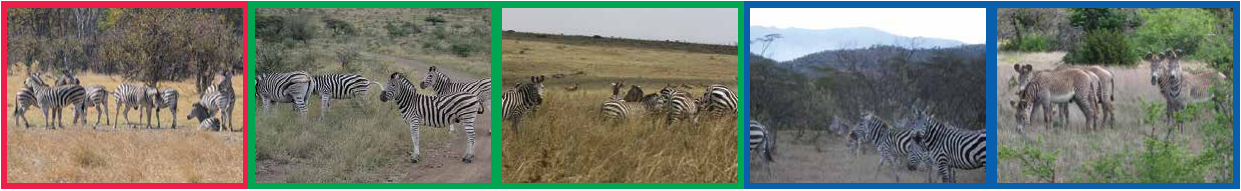}%
    \caption{\small {\bf \ourdataset examples}. 
    The given caption query: ``A herd of zebras standing together in the field''. \textcolor{red}{Red}: original positive. \textcolor{green}{Green}: annotated as ``100\% Yes''. \textcolor{blue}{Blue}: annotated as ``Weak Yes''. More examples are in 
    \cref{subsec:eccv_examples}.
}%
    \label{fig:example_eccv}
\end{figure}%
\begin{figure}[t]
    \centering
    \begin{subfigure}[b]{0.48\textwidth}
        \centering
        \includegraphics[width=\textwidth]{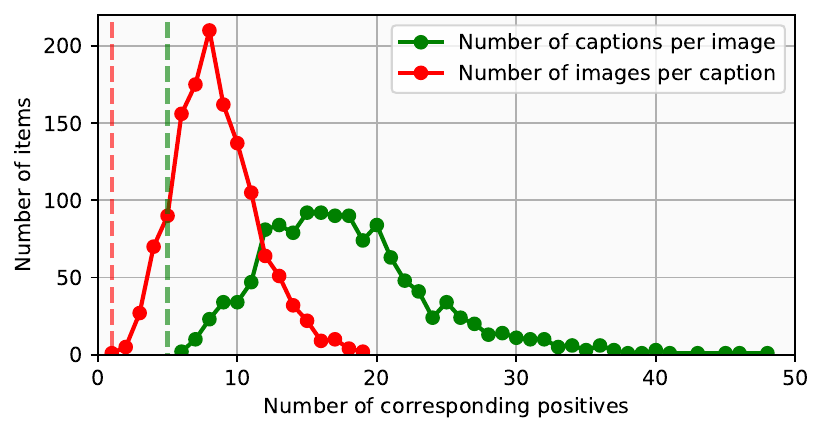}%
        \caption{\small Number of positive pairs.}%
        \label{fig:data_positive_stats}%
    \end{subfigure}%
    \begin{subfigure}[b]{0.48\textwidth}
        \centering
        \includegraphics[width=\textwidth]{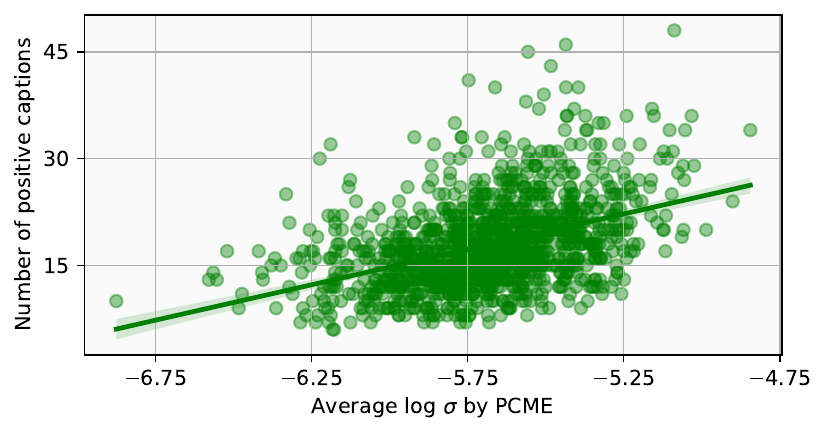}%
        \caption{\small Multiplicity by positive items.}%
        \label{fig:uncertainty_vs_positives}%
    \end{subfigure}%
    \caption{\small {\bf Multiplicity in \ourdataset.} (a) The number of positive pairs in \ourdataset. Dashed lines denote the number of the original COCO positives (1 image for each caption, and 5 captions for each image). \ourdataset contains plenty of positive items per each modality.
    (b) PCME-predicted multiplicity against the number of positive captions for each image.
    There exists a positive correlation.}
    \label{fig:mutliplicity_by_eccv}
\end{figure}

We consider the CxC positives as the additional sixth machine-human verified annotations, and
extend our human-verified positives with CxC positives to construct the final \ourdataset. \cref{tab:positive_pair_statistics} shows the detailed statistics of CxC, human-verified positives, and our \ourdataset. Overall, \ourdataset has $\times$8.47 positive images and $\times$3.58 positive captions than the original dataset. \cref{fig:data_positive_stats} shows the number of positive images and captions per each item; there exist many positives beyond the original COCO associations. We illustrate example image-caption pairs from \ourdataset in \cref{fig:example_eccv} and 
\cref{subsec:eccv_examples}.

We additionally analyze the multiplicity of \ourdataset by PCME \cite{chun2021pcme} that produces a degree of multiplicity (uncertainty) for each query. \cref{fig:uncertainty_vs_positives} shows that more uncertain images correspond to more captions in our dataset. In other words, our new annotations capture the hidden FNs in COCO well.

%% file: tables/models_overview.tex
\begin{table}[t]
\centering
\small
\setlength{\tabcolsep}{3pt}
\setlength{\abovetopsep}{0.5em}
\caption{\small {\bf Overview of the machine annotators.} Differences among five ITM models in terms of architectures and training objectives are shown. ViLT and CLIP are trained on a massive amount of aligned VL data, while other methods only use COCO Caption.}
\label{tab:model_overview}
\begin{tabular}{@{}llll@{}}
\toprule
Model & Text backbone & Visual backbone & Objective function \\ \midrule
PVSE \cite{song2019pvse}  & Bi-GRU \cite{cho2014properties}       & ResNet-152 \cite{resnet} & Multiple instance learning \\
VSRN \cite{Li2019VSRN}  & Bi-GRU        & Faster R-CNN \cite{ren2015faster} & Semantic reasoning matching                \\
PCME \cite{chun2021pcme}  & Bi-GRU        & ResNet-152 & Probabilistic matching \\
ViLT \cite{kim2021vilt}  & \multicolumn{2}{l}{Vision Transformer (ViT-B/32) \cite{vit}} & Vision-language pre-training\\
CLIP \cite{clip}  & Transformer \cite{radford2019language}   & ViT-B/32 & Contrastive learning \\ \bottomrule
\end{tabular}
\end{table}

%% file: tables/dataset_comparisons.tex
\begin{table}[t]
\small
\centering
\setlength{\tabcolsep}{5pt}
\setlength{\abovetopsep}{0.5em}
\caption{\small {\bf Precision and recall of the existing benchmarks measured by our human verified positive pairs.} A low Prec means that many positives are actually negatives, and a low Recall means that there exist many missing positives.}
\label{tab:cxc_pm_pr}
\begin{tabular}{lllll}
\toprule
Dataset & I2T Prec & I2T Recall & T2I Prec & T2I Recall \\ \midrule
Original MS-COCO Caption & 96.9 & 21.1 & 96.4 & 13.8 \\
CxC \cite{parekh2020crisscrossed} & 95.5 & 23.0 & 93.2 & 16.1 \\
Plausible Match ($\zeta=0$) \cite{chun2021pcme} & 65.3 & 56.5 & 56.6 & 61.8 \\ \bottomrule
\end{tabular}
\end{table}

\begin{table}[t]
\small
\centering
\setlength{\tabcolsep}{5pt}
\setlength{\abovetopsep}{0.5em}
\caption{\small {\bf The number of positive images and captions for each dataset.} We show the number of positive items for the subset of the COCO Caption test split. The number of query captions and images are 1,332 and 1,261, respectively.}%
\label{tab:positive_pair_statistics}
\begin{tabular}{@{}lll@{}}
\toprule
Dataset                           & \# positive images    & \# positive captions \\ \midrule
Original MS-COCO Caption          & 1,332                 & 6,305 ($=$1,261$\times$5) \\
CxC \cite{parekh2020crisscrossed} & 1,895 ($\times$1.42)  & 8,906 ($\times$1.41) \\
Human-verified positives          & 10,814 ($\times$8.12) & 16,990 ($\times$2.69) \\
\ourdataset                       & 11,279 ($\times$8.47) & 22,550 ($\times$3.58) \\ \bottomrule
\end{tabular}
\end{table}

%% file: 4_benchmark.tex
\section{Re-evaluation of ITM models on \ourdataset}
\label{sec:benchmark}

In this section, we re-evaluate the existing VL models on our new dataset and previous benchmarks. 
We first introduce the evaluation metrics and comparison methods (\S\ref{subsec:evaluation_metrics_and_comparison_methods}).
We compare the performances and analyze the results (\S\ref{subsec:main_results}).

\subsection{Evaluation metrics and comparison methods}
\label{subsec:evaluation_metrics_and_comparison_methods}

\paragraph{Evalution metrics.}
The existing ITM benchmarks (\eg, COCO Caption) use \recallatk metrics, particularly \recallatone (\rone).
Specifically, previous works measure \rone for 5-fold validation splits (\ie, each split has 1K images), and for the full test split \cite{karpathy2015deep}. The former is called COCO 1K \rk and the latter is called COCO 5K \rk, respectively. Previous studies separately report image-to-text, text-to-image retrieval \rone, \rfive and \rten scores.
However, as shown by Musgrave \etal \cite{musgrave2020metric}, \rk is not an informative metric; embedding spaces with nearly 100\% \rone can have different properties. The problem becomes even worse for the ITM benchmarks, whose queries only have very few (usually only one) references: Even if a model correctly retrieves plausible items that are not among the set of original positives, the current benchmark cannot evaluate the model correctly. It is common to use larger values of $k$ to less penalize wrong yet plausible predictions. However, as shown in \cref{fig:data_positive_stats}, the actual number of plausible positives can be larger than the typical choice of $k$ (\eg, 5 or 10).
Instead, we suggest using \mapr \cite{musgrave2020metric}, a modified \map measured by retrieving $R$ items where $R$ is the number of positives for the query. Previous ITM benchmarks cannot employ \mapr because $R$ is too small (\ie, 1). Thanks to our human-verified ground-truth positives, we can reliably measure \mapr on \ourdataset.

We additionally conduct a human study to confirm that \mapr is more aligned to humans than \rk. We collect 3,200 pairwise preferences of human annotators among
(A) only top-1 is wrong (B) only top-1 is correct (C) top-1 to 5 are wrong but others are correct (D) only top-5 is correct, and (E) all items are wrong. For example, if the number of positives is 8, then (A) shows 0 \rone, 100 \rfive and 66.0 \mapr, (B) shows 100 \rk and 12.5 \mapr, (C) shows 0 \rk and 10.3 \mapr, and (D) shows 0 \rone, 100 \rfive and 2.5 \mapr. We compute user preference scores using Bradley–Terry model \cite{bradley1952rank}. We observe that \mapr is exactly aligned to the human preference score: (A: 70.85, B: 13.15, C: 10.66, D: 4.89, E: 0.44). We provide the details of the human study in 
\cref{subsec:user_study_details}

We also report modified Plausible Match R-Precision (PMRP) scores by changing $R$ to $\min(R, 50)$, because the number of pseudo-positives $R$ can be very large (\eg, larger than 10,000) but most of them are not actual positive (\cref{tab:cxc_pm_pr}).
While Chun \etal \cite{chun2021pcme} proposed to use the average R-Precision for three different thresholds, (\eg, $\zeta = \{0, 1, 2\}$), we only report PMRP when $\zeta = 0$.
We additionally compute \rone, \rfive, and PMRP scores on the original COCO Caption, \rone on CxC, and \rone and R-Precision on \ourdataset to analyze the correlation between each evaluation metric to \eccv \mapr.

\input{tables/main_results}
\input{figures/ranking_figure}
\input{tables/metric_wise_kendalls_tau}

\paragraph{Evaluated methods.}
We compare 25 state-of-the-art VL models, whose trained weights are publicly accessible, categorized into four groups: (1) visual semantic embedding (VSE) methods with the ResNet-152 \cite{resnet} image encoder, and Bi-GRU \cite{cho2014properties} text encoder, including VSE0, VSE++ \cite{faghri2018vsepp}, PVSE \cite{song2019pvse} (K=1 \& K=2), and PCME \cite{chun2021pcme} (the official model and the CutMix pre-trained version); (2) VSE methods with 
region features extracted by Visual Genome \cite{krishna2017visual} pre-trained Faster R-CNN \cite{ren2015faster} based on the implementation by Anderson \etal \cite{Anderson2017up-down} and Lee \etal \cite{lee2018scan}, including VSRN \cite{Li2019VSRN}, VSRN + AOQ \cite{chen2020adaptive}, CVSE \cite{Wang2020CVSE}, SGR, SAF \cite{Diao2021SGRAF}, and VSE$\infty$ with BUTD region, grid and WSL grid features \cite{chen2021vseinfty}\footnote{Techinally speaking, VSE$\infty$ (WSL grid) does not use region features, but CNN features extracted from Instagram-trained ResNext \cite{wslimageseccv2018}. This study treats all VSE$\infty$ variants as region feature-based models for convenience.}. (3) Large-scale VL pre-training (VLP) methods, including pre-trained CLIP with ViT-B/32, ViT-B/16, and ViT/L14 backbones \cite{clip}, pre-trained and fine-tuned ViLT \cite{kim2021vilt}, pre-trained and fine-tuned VinVL \cite{zhang2021vinvl}, and fine-tuned BLIP \cite{li2022blip}. Here, ``pre-trained'' signifies that the model is trained with a massive image-text aligned dataset, but is not specifically trained for COCO Caption; ``fine-tuned'' signifies that the model is fine-tuned on COCO Caption for the ITM task.
We note that VL transformers except CLIP need $O(|C| \times |I|)$ forward operations to compute the full pairwise ranks between $|C|$ number of captions and $|I|$ number of images, while other methods only need $O(|I|) + O(|C|)$ forward operations to compute the full pairwise ranks based on the cosine similarity. For example, VinVL takes 25 hours to compute the full pairwise ranks for the COCO Caption test split by a single A100 GPU core, while VSE++ only takes 1 minute in the same environment.
(4) PVSE models with different negative mining (NM) methods, including no NM, semi-hard NM (SHM) \cite{schroff2015facenet}, and hardest NM (HNM) \cite{faghri2018vsepp}.

We use the official trained weights for each model with a few exceptions. We re-implement VSE0, VSE++, PCME with CutMix pre-trained ResNet, and PVSE models with various NM strategies.
The training details are in 
\cref{subsec:training_details}.

\subsection{Re-evaluation of ITM methods}
\label{subsec:main_results}

\cref{tab:main_results} and \cref{fig:bump_main} shows the full comparisons of 25 VL models with different evaluation metrics.
We report the Kendall's rank correlations (tau-b) between metrics in \cref{tab:metric_wise_kendall_full}; larger $\tau$ denotes two metrics are more correlated.
We report the full table including modality-wise results, \rfive and \rten scores in 
\cref{subsec:more_evaluation_full_table}.
We first observe that \rk scores across different datasets have high correlations among themselves (\cref{fig:bump_main_b} and 
\cref{subsec:more_evaluation_full_table}).
).
In terms of the ranking correlation, we observe that COCO 1K R@1 shows almost $\tau$=0.9 with the ranking yielded by R@5 (0.87), COCO 5K R@1 (0.89) and R@5 (0.97), or CxC R@1 (0.89). This implies that measuring \recallatk on different benchmarks, such as the original COCO Caption, CxC, and \ourdataset are not more informative than only measuring \recallatk on COCO 1K or 5K.
On the other hand, the rankings by COCO 1K are not preserved well to PMRP (0.45), \eccv \rone (0.72), \eccv R-Precision (0.39) and \eccv \mapr (0.47) in Kendall's $\tau$.
This implies that enlarging $K$ of \rk (\eg, using \rfive, \rten instead of \rone) cannot be an alternative of \mapr because \rk metrics are highly correlated each other as shown in \cref{tab:metric_wise_kendall_full}.
We also observe that the rankings by PMRP are relatively less correlated to the other metrics, such as COCO \rone (0.45), \eccv \rone (0.29) or \eccv \mapr (0.20) in Kendall's $\tau$.

Our re-evaluation shows that existing ITM evaluation benchmarks can overestimate the VL model performance by focusing only on COCO \rone, where the rankings between COCO \rone and \eccv \mapr are not largely preserved. 
For example, we observe that the hardest negative mining technique \cite{faghri2018vsepp}, previously deemed useful for ITM tasks, is actually selectively effective for \rone, rather than for the actual task itself. Under our new metrics like \eccv \mapr, we observe that the milder strategy of semi-hard negative mining is more effective -- See \cref{fig:comp_groups_b}. Chun \etal \cite{chun2021pcme} also observed a similar pattern in the CUB Caption dataset \cite{cub} by using the class-defined positives. Our finding is the first observation in the practical large-scale VL dataset. 
Similarly, we observe that many large-scale VL pre-training methods with high \rone scores show inferior \eccv \mapr scores compared to other visual semantic embedding techniques. For example, CLIP ViT-L/14 shows superior COCO 1K \rone than PCME (55.4\% and 40.1\%, respectively). However, in terms of \eccv \mapr, CLIP shows inferior performances than PCME (28.0\% and 37.1\%, respectively).

Similarly, we observe that PMRP shows different behaviors compared to other metrics.
Especially, we observe that the contrastive models without a negative mining strategy are specialized to PMRP metric -- \cref{fig:comp_groups_a}. We presume that it is because the contrastive learning strategy enforces the features with similar objects to be mapped to a similar embedding space. 
In contrastive the best models on COCO and \eccv (\eg, BLIP, VinVL, and VSE$\infty$) show inferior PMRP scores -- \cref{fig:comp_groups_c}. We presume that it is because PMRP only captures the existence or absence of the objects, while an optimal retrieval also should consider the plausibility between matched image-caption pairs.

\input{figures/modelwise_ranking_figure}

%% file: tables/main_results.tex
\begin{table}[t!]
\small
\centering
\setlength{\tabcolsep}{4pt}
\setlength{\abovetopsep}{0.5em}
\caption{\small {\bf Re-evaluating VL models.} ECCV Caption mAP@R, R-Precision (R-P), Recall@1 (\rone), CxC \rone, COCO 1K \rone, 5K \rone, PMRP, and RSUM (the summation of COCO 1K recalls) are shown. The numbers are the average between the image-to-text retrieval and text-to-image retrieval results. Full numbers for each modality and COCO \rfive, \rten results are in Appendix D.3. $^\dagger$ denotes our re-implemention and ``zero-shot'' for VinVL and ViLT denotes VL pre-trained models without fine-tuning on the COCO Caption for the retrieval task.}
\label{tab:main_results}
\resizebox{\columnwidth}{!} {
\begin{tabular}{@{}lcccccccccc@{}}
\toprule
& \multicolumn{3}{c}{ECCV Caption} &  & CxC &  & \multicolumn{4}{c}{COCO} \\
& mAP@R & R-P & R@1 &  & R@1 &  & 1K R@1 & 5K R@1 & PMRP & RSUM \\ \midrule
\multicolumn{11}{c}{ResNet-152 \cite{resnet} image encoder + Bi-GRU \cite{cho2014properties} text encoder} \\ \midrule
VSE0$^\dagger$ \cite{faghri2018vsepp} & 22.67 & 33.27 & 55.55 &  & 24.24 &  & 44.23 & 22.27 & 46.95 & 418.9 \\
VSE++$^\dagger$ \cite{faghri2018vsepp} & 35.01 & 45.50 & 73.11 &  & 37.95 &  & 59.81 & 35.79 & 54.26 & 483.5 \\
PVSE K=1 \cite{song2019pvse} & 33.98 & 44.49 & 73.25 &  & 38.38 &  & 60.10 & 36.20 & 53.56 & 483.9 \\
PVSE K=2 \cite{song2019pvse} & \underline{40.26} & \underline{49.92} & \underline{76.74} &  & \underline{40.18} &  & 61.67 & \underline{38.13} & 55.52 & 490.4 \\
PCME \cite{chun2021pcme} & 37.11 & 47.82 & 74.79 &  & 40.09 &  & \underline{61.72} & 38.03 & \underline{56.71} & \underline{490.6} \\
PCME {\scriptsize (CutMix \cite{cutmix} pre-trained)}$^\dagger$ & \textbf{41.74} & \textbf{51.45} & \textbf{78.67} &  & \textbf{41.70} &  & \textbf{62.71} & \textbf{39.51} & \textbf{57.65} & \textbf{495.3} \\ \midrule
\multicolumn{11}{c}{Region features based on Bottom-up Attention \cite{Anderson2017up-down} and SCAN \cite{lee2018scan}} \\ \midrule
VSRN \cite{Li2019VSRN} & \textbf{42.28} & \textbf{51.84} & 81.51 &  & 48.85 &  & 69.49 & 46.74 & 55.44 & 515.9 \\
VSRN + AOQ \cite{chen2020adaptive} & 40.94 & 50.65 & 81.53 &  & 50.10 &  & 70.48 & 48.14 & 56.41 & 520.7 \\
CVSE \cite{Wang2020CVSE} & 37.35 & 47.51 & 76.70 &  & 45.82 &  & 67.01 & 43.80 & 56.49 & 511.1 \\
SGR \cite{Diao2021SGRAF} & 35.80 & 46.04 & 78.77 &  & 50.60 &  & 69.66 & 48.86 & 56.91 & 518.0 \\
SAF \cite{Diao2021SGRAF} & 35.96 & 46.19 & 78.36 &  & 49.58 &  & 70.15 & 47.80 & \underline{57.21} & 518.8 \\
VSE$\infty$ (BUTD region) \cite{chen2021vseinfty} & 40.46 & 49.97 & 82.52 &  & 52.40 &  & 72.22 & 50.38 & 56.64 & 526.8 \\
VSE$\infty$ (BUTD grid) \cite{chen2021vseinfty} & 40.40 & 50.09 & \underline{83.01} &  & \underline{53.47} &  & \underline{73.42} & \underline{51.60} & 56.87 & \underline{530.9} \\
VSE$\infty$ (WSL grid) \cite{chen2021vseinfty} & \underline{42.41} & \underline{51.43} & \textbf{86.44} &  & \textbf{60.79} &  & \textbf{78.27} & \textbf{59.01} & \textbf{57.65} & \textbf{545.1} \\ \midrule
\multicolumn{11}{c}{Large-scale Vision-Language pre-training} \\ \midrule
CLIP ViT-B/32 \cite{clip} & 26.75 & 36.91 & 67.08 &  & 41.97 &  & 59.47 & 40.28 & 55.32 & 471.9 \\
CLIP ViT-B/16 \cite{clip} & 29.25 & 38.99 & 71.05 &  & 44.26 &  & 62.02 & 42.69 & 56.58 & 481.0 \\
CLIP ViT-L/14 \cite{clip} & 27.98 & 37.80 & 72.17 &  & 48.14 &  & 64.83 & 46.44 & \textbf{57.70} & 491.5 \\
VinVL (zero-shot) \cite{zhang2021vinvl} & 22.18 & 32.93 & 55.19 &  & 33.74 &  & 54.11 & 32.07 & 47.26 & 457.0 \\
VinVL \cite{zhang2021vinvl} & \textbf{40.81} & \textbf{49.55} & \underline{87.77} & & \underline{67.76} & & \underline{82.38} & \underline{66.39} & 54.72 & \underline{555.5} \\
ViLT (zero-shot) \cite{kim2021vilt} & 26.84 & 36.81 & 69.00 &  & 50.35 &  & 69.68 & 48.63 & 57.38 & 519.3 \\
ViLT \cite{kim2021vilt} & 34.58 & 44.27 & 77.81 &  & 53.72 &  & 72.75 & 52.18 & \underline{57.63} & 528.6 \\
BLIP \cite{li2022blip} & \underline{40.52} & \underline{48.43} & \textbf{90.99} &  & \textbf{74.30} &  & \textbf{86.12} & \textbf{73.11} & 57.17 & \textbf{564.4} \\ \midrule
\multicolumn{11}{c}{Different negative mining (NM) strategies} \\ \midrule
PVSE K=1, {\scriptsize No NM}$^\dagger$ & 33.34 & 44.44 & 67.99 &  & 32.69 &  & 54.86 & 30.65 & \textbf{56.67} & 469.3 \\
PVSE K=1, {\scriptsize Semi-hard NM}$^\dagger$ \cite{schroff2015facenet} & \textbf{36.63} & \textbf{47.36} & \textbf{73.97} &  & 38.17 &  & 59.85 & 36.00 & 55.15 & 485.1 \\
PVSE K=1, {\scriptsize Hardest NM}$^\dagger$ \cite{faghri2018vsepp} & 35.76 & 46.50 & 73.68 &  & \textbf{39.02} &  & \textbf{60.60} & \textbf{36.88} & 54.37 & \textbf{486.9} \\ \bottomrule
\end{tabular}
}
\end{table}

%% file: figures/ranking_figure.tex
\begin{figure}[t!]
    \centering
    \begin{subfigure}[b]{\linewidth}
        \centering
        \includegraphics[width=\textwidth]{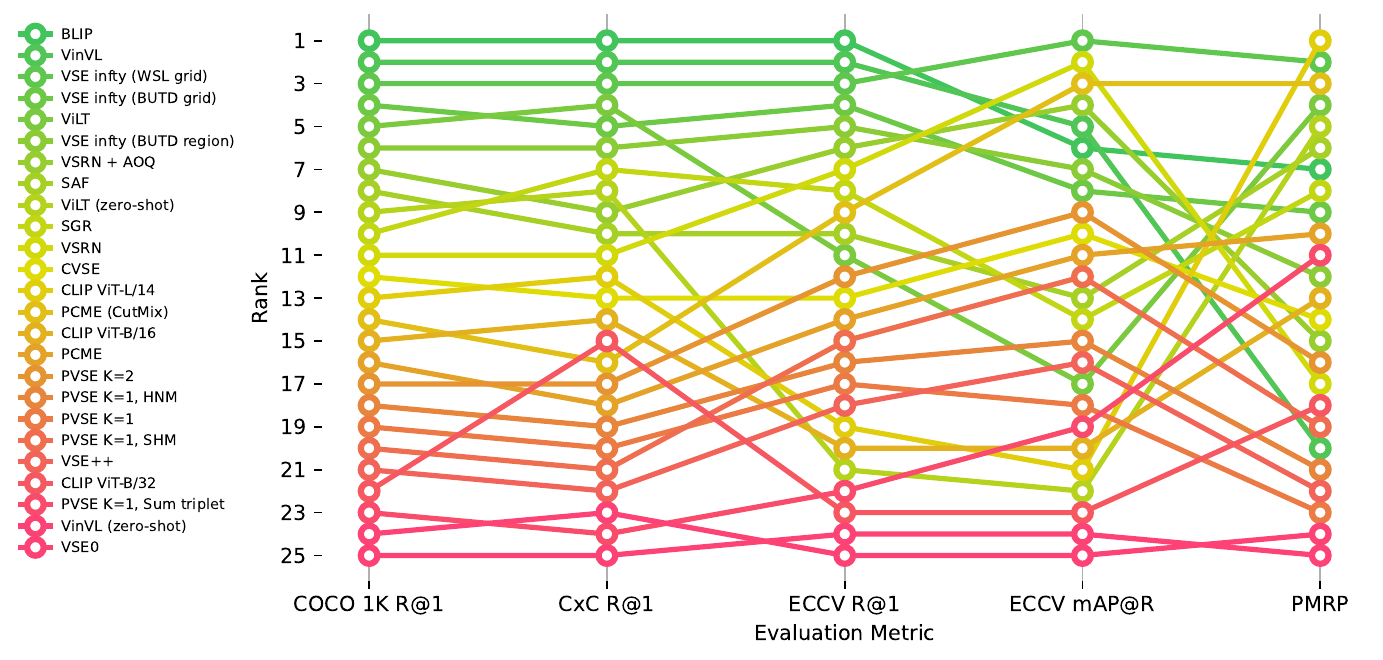}
        \caption{\small Comparison of COCO, CxC, ECCV and PMRP.}
        \label{fig:bump_main_a}
    \end{subfigure}
    \hfill
    \begin{subfigure}[b]{0.57\linewidth}
        \centering
        \includegraphics[width=\textwidth]{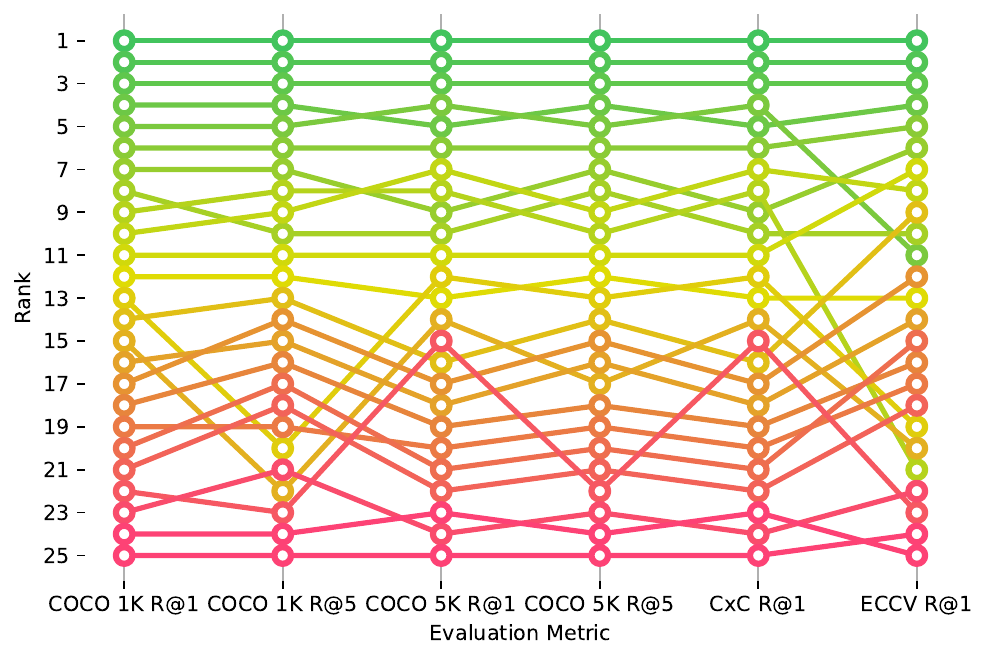}
        \caption{\small Comparison of Recall@1 metrics.}
        \label{fig:bump_main_b}
    \end{subfigure}%
    \hfill
    \begin{subfigure}[b]{0.42\linewidth}
        \centering
        \includegraphics[width=\textwidth]{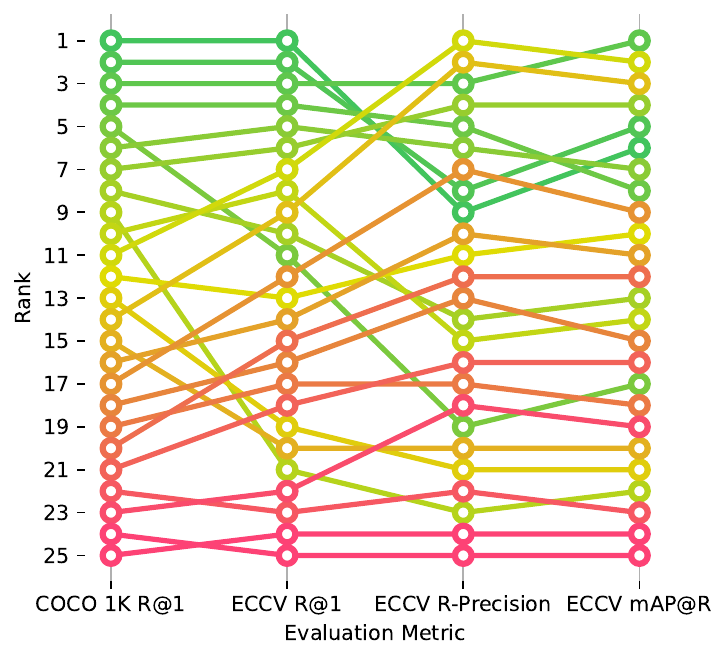}
        \caption{\small Comparison of ECCV metrics.}
        \label{fig:bump_main_c}
    \end{subfigure}
    \caption{\small {\bf Ranking correlation between different evaluation metrics.} Ranking of methods is largely perserved between COCO and CxC Recall@1, while it is rarely preserved among COCO Recall@1, ECCV mAP@R and PMRP.}
    \label{fig:bump_main}
\end{figure}

%% file: tables/metric_wise_kendalls_tau.tex
\begin{table}[t]
\small
\centering
\setlength{\tabcolsep}{6pt}
\setlength{\abovetopsep}{0.5em}
\caption{\small {\bf Rank correlations between evaluation metrics.} Higher $\tau$ denotes two rankings are highly correlated, while $\tau$ values near zero denotes two rankings are barely correlated. We highlight the highly correlated pairs ($\tau > 0.8$) with \hc{red} text. ``RSUM'' denotes the summation of COCO 1K R@1s, R@5s, R@10s for each modality.}
\label{tab:metric_wise_kendall_full}
\resizebox{\columnwidth}{!} {
\begin{tabular}{@{}lcccccccccccccccc@{}}
\toprule
& \multicolumn{4}{c}{COCO 1K} && \multicolumn{3}{c}{COCO 5K} && CxC && \multicolumn{3}{c}{ECCV} && \multicolumn{1}{c}{COCO} \\
& R@1 & R@5 & R@10 & RSUM & & R@1 & R@5 & R@10 &  & R@1 &  & R@1 & R-P & \mapr &  & PMRP \\ \midrule
COCO 1K R@1 & - & \hc{0.87} & \hc{0.86} & \hc{0.94}& & \hc{0.89} & \hc{0.97} & \hc{0.92}& & \hc{0.89}& & 0.72 & 0.39 & 0.47& & 0.45 \\
COCO 1K R@5 & \hc{0.87} & - & \hc{0.97} & \hc{0.93}& & 0.79 & \hc{0.88} & \hc{0.93}& & 0.79& & \hc{0.81} & 0.49 & 0.58& & 0.39 \\
COCO 1K R@10 & \hc{0.86} & \hc{0.97} & - & \hc{0.92}& & 0.77 & \hc{0.86} & \hc{0.91}& & 0.77& & 0.79 & 0.49 & 0.57& & 0.43 \\
RSUM & \hc{0.94} & \hc{0.93} & \hc{0.92} & -& & \hc{0.84} & \hc{0.94} & \hc{0.95}& & \hc{0.84}& & 0.77 & 0.43 & 0.52& & 0.43 \\
COCO 5K R@1 & \hc{0.89} & 0.79 & 0.77 & \hc{0.84}& & - & \hc{0.89} & \hc{0.83}& & \hc{1.00}& & 0.65 & 0.30 & 0.39& & 0.45 \\
COCO 5K R@5 & \hc{0.97} & \hc{0.88} & \hc{0.86} & \hc{0.94}& & \hc{0.89} & - & \hc{0.95}& & \hc{0.89}& & 0.75 & 0.41 & 0.50& & 0.43 \\
COCO 5K R@10 & \hc{0.92} & \hc{0.93} & \hc{0.91} & \hc{0.95}& & \hc{0.83} & \hc{0.95} & -& & \hc{0.83}& & 0.80 & 0.47 & 0.55& & 0.38 \\
CxC R@1 & \hc{0.89} & 0.79 & 0.77 & \hc{0.84}& & \hc{1.00} & \hc{0.89} & \hc{0.83}& & -& & 0.65 & 0.30 & 0.39& & 0.45 \\
ECCV R@1 & 0.72 & \hc{0.81} & 0.79 & 0.77& & 0.65 & 0.75 & 0.80& & 0.65& & - & 0.65 & 0.74& & 0.29 \\
ECCV R-Precision & 0.39 & 0.49 & 0.49 & 0.43& & 0.30 & 0.41 & 0.47& & 0.30& & 0.65 & - & \hc{0.90}& & 0.17 \\
ECCV mAP@R & 0.47 & 0.58 & 0.57 & 0.52& & 0.39 & 0.50 & 0.55& & 0.39& & 0.74 & \hc{0.90} & -& & 0.20 \\
PMRP & 0.45 & 0.39 & 0.43 & 0.43& & 0.45 & 0.43 & 0.38& & 0.45& & 0.29 & 0.17 & 0.20& & - \\
\bottomrule
\end{tabular}
}
\end{table}

%% file: figures/modelwise_ranking_figure.tex
\begin{figure}[t!]
    \centering
    \begin{subfigure}[b]{0.33\linewidth}
        \centering
        \includegraphics[width=\textwidth]{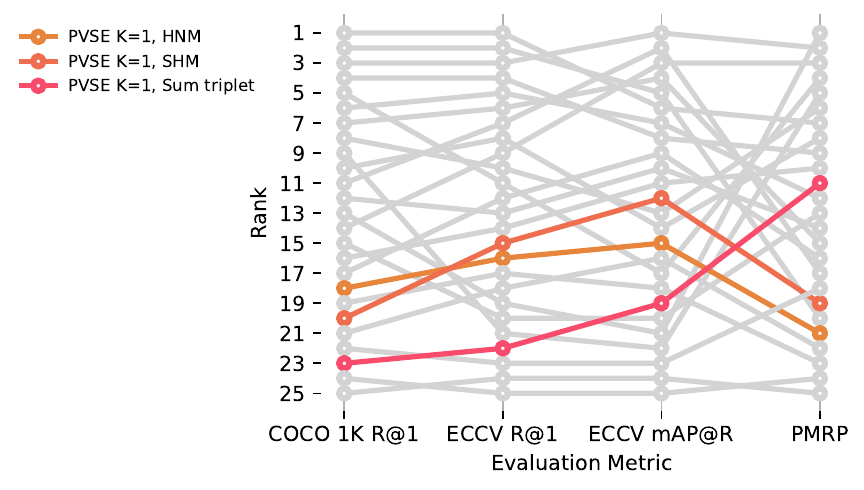}
        \caption{\small Triplet mining strategies.}
        \label{fig:comp_groups_b}
    \end{subfigure}%
    \hfill
    \begin{subfigure}[b]{0.33\linewidth}
        \centering
        \includegraphics[width=\textwidth]{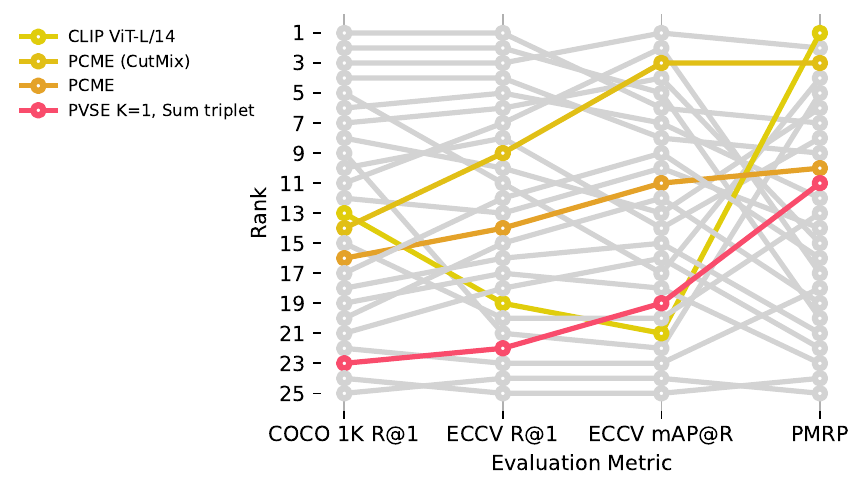}
        \caption{\small Contrastive methods.}
        \label{fig:comp_groups_a}
    \end{subfigure}%
    \hfill
    \begin{subfigure}[b]{0.33\linewidth}
        \centering
        \includegraphics[width=\textwidth]{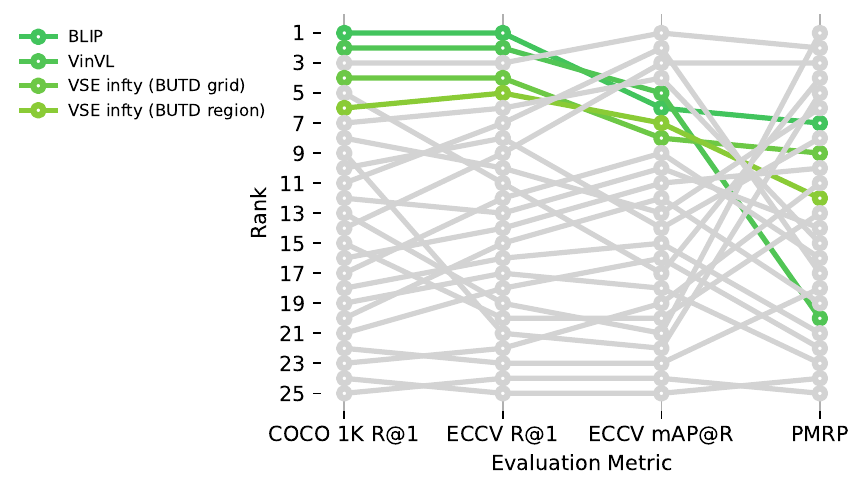}
        \caption{\small Best \rone models.}
        \label{fig:comp_groups_c}
    \end{subfigure}%
    \caption{\small {\bf Rankings of different VL models.} Ranking of (a) PVSE models with diverse triplet mining strategies (b) contrastive methods (c) the best models are shown.}
    \label{fig:comp_groups}
\end{figure}

%% file: 5_discussion.tex
\section{Discussion and Limitations}
\label{sec:discussion}

\paragraph{Potential machine biases in our dataset.}
Our dataset construction process contains the MITL annotation process, where the choice of machine annotators can potentially harm the dataset quality.
The positives in our dataset are the retrieved items by the machine annotators. If the machines are biased towards undesired patterns (\eg favoring certain items over the others), future methods built on our benchmark will overfit those patterns. In this work, we employ five diverse machine annotators to reduce the potential biases by models. 
In \cref{sec:bias_analysis},
we explore and quantify the effect of the choice of multiple machine annotators on the dataset quality. From the study, we can conclude that our strategy (using more models) is effective to mitigate biases by a specific model.

\paragraph{Scale of \ourdataset.}
In this work, we subsample 1,333 caption queries (5.3\% of the full caption queries) and 1,261 image queries (25.2\% of the full image queries) to reduce the scale of annotations. Note that without subsampling, we need to verify (25,000 + 5,000) $\times$ 5 $\times$ 5 = 750K pairs, which costs 16 times more than our current version, almost \$60K. Because we only subsample queries, not limiting the gallery samples, our dataset is an unbiased subset of the original COCO Caption.
To scale up \ourdataset, we have to reduce the human verification costs by reducing the total number of human verification. This can be achievable by applying a multi-turn MITL annotation process that alternatively repeats training machine annotators with human-annotated associations and verifying machine annotations by human workers.
After enough iterations of the multi-turn MITL annotation process, we can automatically scale up our annotations by using the high-quality machine annotators while only low confident associations are verified by humans.

\paragraph{Noisy annotations.}
Despite our additional verification process to keep the quality of the annotations, there can be noisy annotations (\ie, false positives) in \ourdataset due to the noisy nature of crowdsourcing annotations. The noisy annotations can also occur because we use both ``100\% YES'' and ``Partially YES'' to build positive pairs.
However, we still encourage to use \ourdataset for evaluating VL models, because the existing datasets are noisier; they usually have only one positive item per each query and they have tremendously many FNs. On the other hand, noisy annotations of our dataset are still ``plausible'' rather than ``wrong''.
We provide more discussion in 
\cref{sec:noisy_annotations}.
Finally, we expect that a multi-turn MITL process can improve not only the labeling cost but also the annotation quality as shown by Benenson \etal \cite{OpenImagesV5}.

%% file: 6_conclusion.tex
\section{Conclusion}
MS-COCO Caption is a popular dataset for evaluating image-text matching (ITM) methods.
Despite its popularity, it suffers from a large number of missing positive matches between images and captions.
Fully annotating the missing positives with human labor incurs prohibitive costs.
We thus rely on machine annotators to propose candidate positive matches and let crowdsourced human annotators verify the matches.
The resulting ITM evaluation benchmark, \ourdatasetfull, contains $\times$8.47 positive images and $\times$3.58 positive captions compared to the original MS-COCO Caption.
We have re-evaluated 25 ITM methods on \ourdataset with \mapr, resulting in certain changes in the ranking of methods.
We encourage future studies on ITM to evaluate their models on \eccv \mapr that not only focuses on the correctness but also on the diversity of top-$k$ retrieved items.

%% file: 7_acknowledgement.tex
\section*{Author Contributions}

Main project idea (\ie, false negative problem) is from S Chun and his previous work \cite{chun2021pcme}.
S Chun led the project; the other authors actively and significantly contributed to the project with advice and feedback.
S Chun, M Chang, W Kim and SJ Oh jointly designed the MITL annotation process; especially, M Chang and W Kim significantly contributed to the design of HIT based on the Human-centered design approach.
W Kim conducted large-scale VL transformers (\eg, ViLT, VinVL) retrieval experiments for annotation and evaluation.
S Chun implemented and conducted the MITL annotation pipeline and the evaluation pipeline.
S Park contributed to HIT verification, the final data cleanup and the data construction.
S Chun performed and interpreted data analysis and evaluation.
M Chang helped to analyze the HIT results from the HCI point of view.
S Chun and SJ Oh wrote the initial version of the manuscript. All authors contributed to the final manuscript.

%% file: 99_appendix.tex
\appendix
\numberwithin{equation}{section}
\numberwithin{figure}{section}
\numberwithin{table}{section}
We include additional materials in this document. 
We first describe the details of our machine annotators (\cref{sec:machine_annotators}), including the explanation of each model (\cref{subsec:machine_annotators_detail}) and diversity between each model (\cref{subsec:machine_annotators_kendall}). We provide the details of Human Intelligence Tasks (HITs) for \ourdataset construction (\cref{sec:mturk_hits}), such as detailed questionnaire (\cref{subsec:mturk_hits_detail}), MTurk worker statistics (\cref{subsec:mturk_annotators}) and the results (\cref{subsec:mturk_results}). \cref{sec:eccv_postprocessing} describes the post-processing details, including the full list of invalid items (\cref{subsec:invalid_items}) and the examples of \ourdataset (\cref{subsec:eccv_examples}).
We include more evaluation results in \cref{sec:more_evaluation}, such as user study details for comparing \mapr and \recallatk (\cref{subsec:user_study_details}), training details of re-implemented methods (\cref{subsec:training_details}), the full results with various evaluation metrics (\cref{subsec:more_evaluation_full_table}).
Finally, we provide the full bias analysis in \cref{sec:bias_analysis} and the discussions of noisy crowdsource annotations in \cref{sec:noisy_annotations}.

\section{\ourdataset Machine Annotators Details}
\label{sec:machine_annotators}

\subsection{Machine annotators}
\label{subsec:machine_annotators_detail}
To cover both diversity and practical relevance, we have choose five state-of-the-art cross-modal retrieval models with diverse properties.
\begin{itemize}
    \item[\textbullet] VSRN \cite{Li2019VSRN} builds up connections between image regions, and perform reasoning with Graph Convolutional Networks to generates features with semantic relationships. VSRN uses the Faster R-CNN detector \cite{ren2015faster} as the visual encoder following \cite{Anderson2017up-down}. VSRN employs the triplet loss \cite{schroff2015facenet} with hardest negative mining (HNM) \cite{faghri2018vsepp}. 
    \item[\textbullet] PVSE \cite{song2019pvse} learns a one-to-many function to solve ambiguous matching by one-to-one function. PVSE is a multi-headed model, focusing on diverse matching between two diverse concepts. PVSE also employs the triplet loss with HNM as VSRN.
    \item[\textbullet] PCME \cite{chun2021pcme} is a stochastic model for learning many-to-many correspondences in multi-modal matching tasks. PCME is trained by a probabilistic matching objective function based on the pair-wise matching loss.
    \item[\textbullet] ViLT \cite{kim2021vilt} is a vision-language pre-training method with massive paired data (4.1M images and 9.9M captions). While other methods have separated text and visual backbones, ViLT has a unified shared Transformer \cite{vaswani2017attention} backbone for text and visual modalities.
    \item[\textbullet] CLIP \cite{clip} is a contrastive approach for massive but noisy associations and shows powerful zero-shot classification performances. CLIP is trained with 4M image and caption pairs. We use the ViT-B/32 CLIP, the largest one when we start the annotation process.
\end{itemize}

\begin{figure}[ht]
    \centering
    \includegraphics[width=\linewidth]{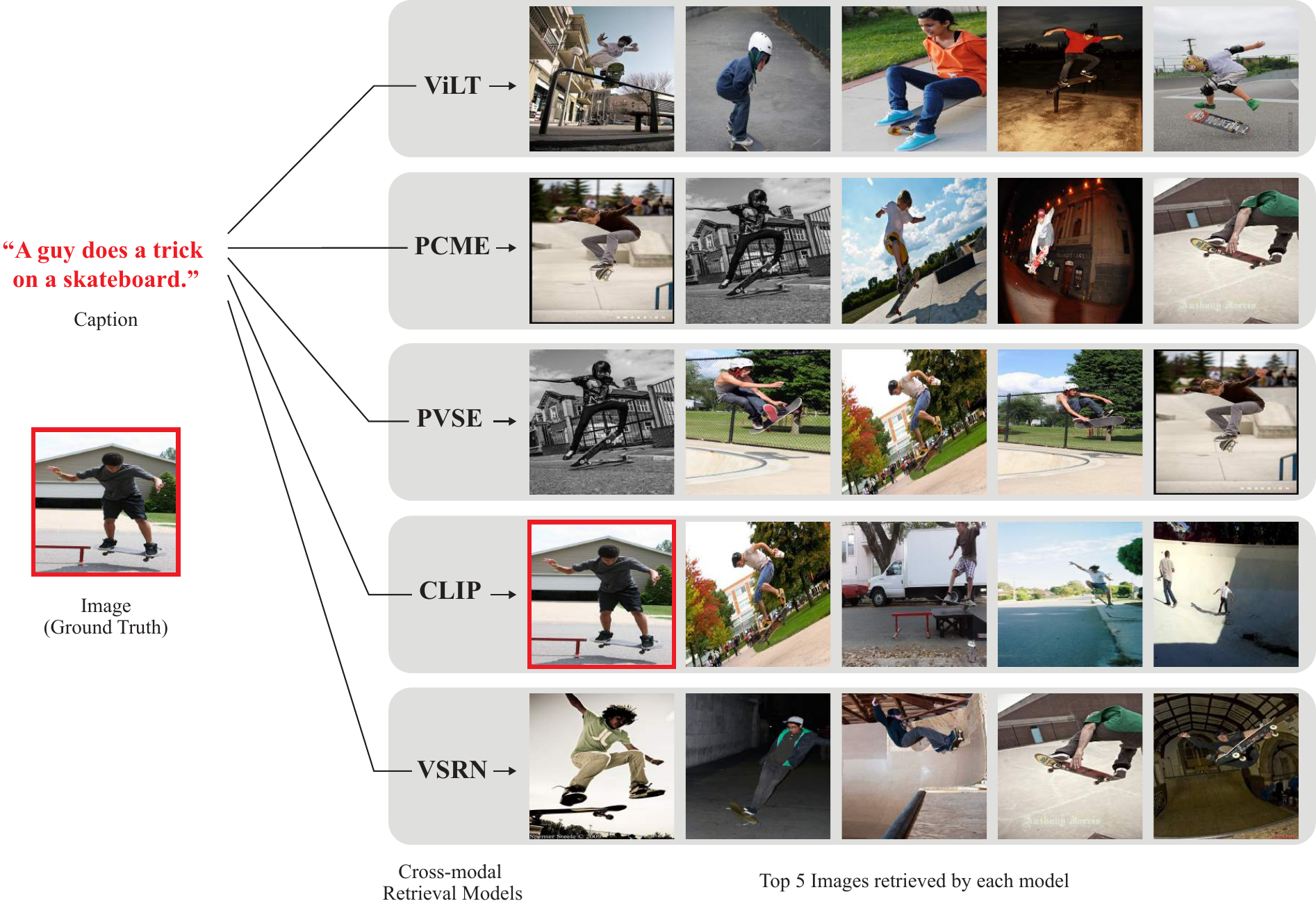}
    \caption{\small {\bf Example retrieved images by the machine annotators.}  For the given caption (``A guy does a trick on a skateboard.''), we show the top-5 images retrieved by models. The matched pair in the dataset is denoted by red boxes.}
    \label{fig:coco_retrieval_results}
\end{figure}

PVSE, VSRN and PCME use pre-trained visual backbones (ImageNet-trained ResNet, Visual Genome \cite{krishna2017visual}-trained Faster R-CNN) and only use COCO Caption dataset as the training dataset.
We use the official weights provided by the authors, except for PCME. We re-train PCME with CutMix \cite{cutmix} pre-trained ResNet-152. This slightly boosts the original performances.
We illustrate the example retrieved images by each model in \cref{fig:coco_retrieval_results}.

\subsection{Diversity between machine annotators}
\label{subsec:machine_annotators_kendall}

\input{tables/model_kendall}

We illustrate the quantify of the diversity between machine annotators by their retrieved items in \cref{tab:model_diff}.
We retrieve 25 images by each model from the COCO validation captions, and measure (1) Kendall's $\tau$ (\cref{tab:model_kendall}), and (2) the average ranking of the top-1 retrieved items by a model of another model.

\cref{tab:model_kendall} shows the Kendall's rank correlation coefficients (Kendall's $\tau$) between the models.
Kendall's $\tau$ is computed on two ranked lists $[x_1, x_2, \ldots, x_n]$ and $[y_1, y_2, \ldots, y_n]$. We say that two pairs ,$(x_i, x_j)$ and $(y_i, y_j)$, \textit{agree} if either ($x_i > x_j$ and $y_i > y_j$) or ($x_i < x_j$ and $y_i < y_j$). Kendall's $\tau$ is computed by $\tau = \frac{\text{\#\ignorespaces agreed pairs} - \text{\#\ignorespaces non-agreed pairs}}{\text{\#\ignorespaces all pairs}}$. We use the tau-B variant for the tie-breaking.

\section{Human Intelligence Tasks (HITs) for \ourdataset Construction}
\label{sec:mturk_hits}

\subsection{HIT details}
\label{subsec:mturk_hits_detail}

\begin{figure}[ht!]
    \centering
    \includegraphics[width=\linewidth]{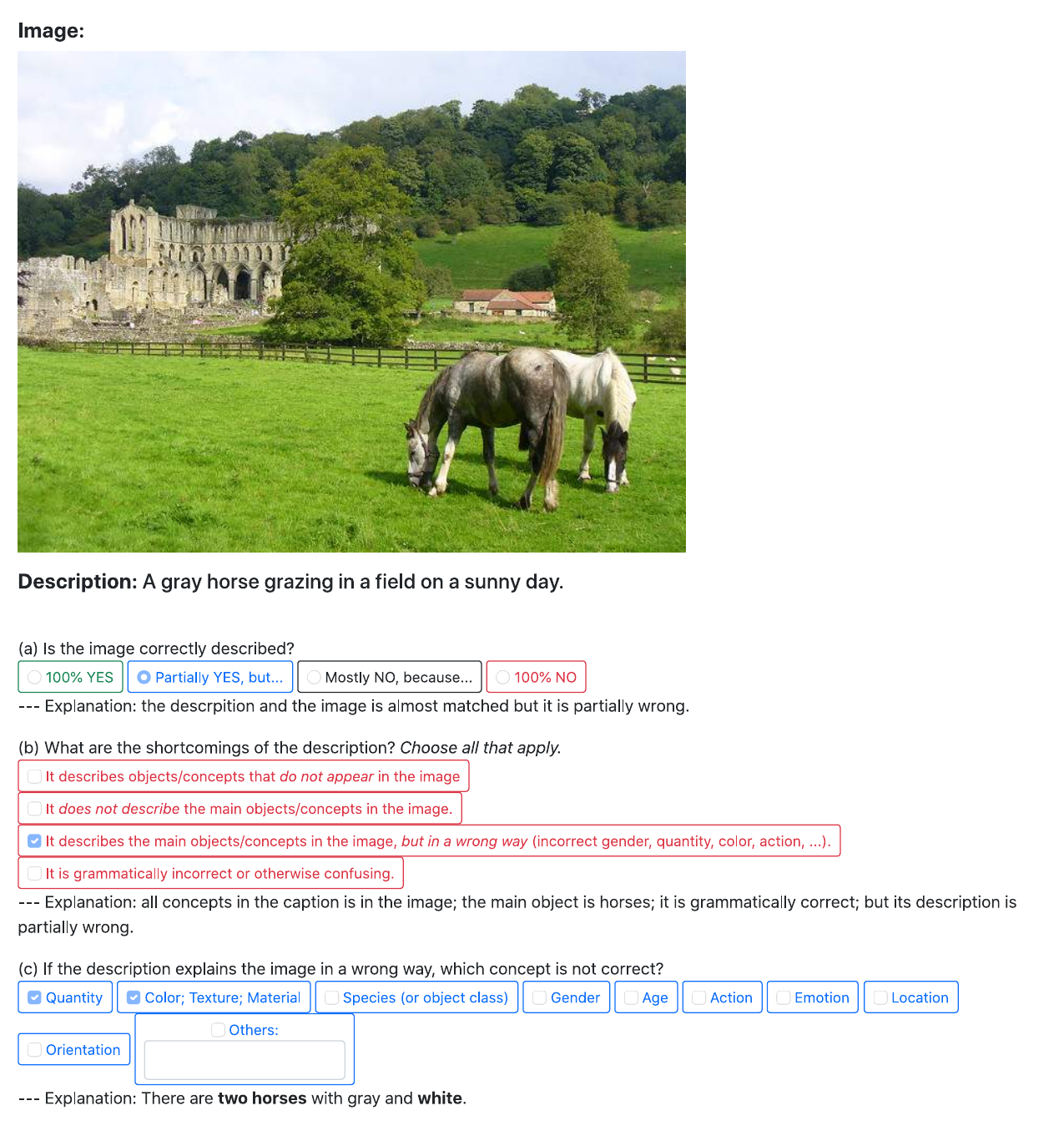}
    \caption{\textbf{Example question in a MTurk HIT.} The question asks whether the image is correctly described. If unsure (``Partially YES, but ...'' or ``Mostly NO, because ...''), the question prompts the worker to provide the rationale. There are 20 of such questions in each HIT.}
    \label{fig:mturk_example}
\end{figure}

The example HIT for crowd workers is shown in \cref{fig:mturk_example}. Each of the 20 questions in the HIT ask the workers to select the degree of belief that the given image-description pair is a positive match. 
We have designed the HITs in such a way that not only the positivity of the match is recorded, but also the degrees and rationales for the workers' judgments are collected.
Workers can choose among ``100\% YES'', ``Partially YES, but'', ``Mostly NO, because'', and ``100\% NO''. Here, we use four choices instead three level (``YES'', ``Not Sure'', and ``NO'') to avoid encouraging the workers to select ``Not Sure'' for all questions. 
If a worker chooses ``Partially YES, but'' or ``Mostly NO, because'', then they are asked further questions on the rationale behind their uncertainty. Four possible shortcomings for the image-description match are presented as choices: ``the description describes concepts that \textit{do not appear} in the image'', ``the description \textit{does not describe} the main concepts in the image'', ``the description describes the main concepts in \textit{a wrong way}'', and ``the description is grammatically incorrect''. Finally, if a worker thinks the description describes the image in a wrong way, we ask \textit{how} the description is wrong. The possible choices here (\eg, quantity, color, \ldots) have been crystallized from an internal, preliminary study.

We make two separated HITs for the annotation process. In the first stage, we verify the results of the image-to-caption retrieval results of five models. We also ask the crowd workers to justify their answer if they choose ``Partially YES'' or ``Mostly NO''. In the second stage, we verify the results of the caption-to-image retrieval results of the models. After we analyzed the first HITs, we have concluded that the justification stage is not highly useful as our expectation. We omit the justification questions in the second stage for reducing annotation costs. The first annotation round was between 23rd Aug 2021 to 7th Sep 2021. During the first stage, 1,000 HITs are verified by human annotators. The second stage was between 24th Jun 2022 to 10th Feb 2022, and 1,160 HITs are verified during this stage.

\input{tables/worker_stats}

\subsection{MTurk workers}
\label{subsec:mturk_annotators}

Before launching the crowdsourcing on AMT, we have conducted an in-lab study involving 70 HITs and 27 workers for 4 days. We have observed that if workers continuously complete HITs, the average elapsed time per HIT is about 4 to 8 minutes. Based on this estimate, we have set the compensation level for each HIT to \$1.4 so that a worker can earn \$15 per hour in the first stage. For the second stage, we have set the the compensation level for each HIT to \$0.65, based on the similar in-lab study without justification questions. The final costs including platform fees for the first and the second stages are \$1.65 and \$0.78, respectively.

In the main crowdsourcing phases, crowd workers are recruited through AMT. The detailed statistics for workers are shown in \cref{tab:worker_stats}. Overall, 970 unique workers have completed 2,969 unique Human Intelligence Tasks (HITs), while 807 HITs of them (37.3\%) have been rejected by our qualification process.
The average elapsed time for each HIT of the first and second annotation phase are 9.5 and 13.8 minutes, respectively. The average HITs per worker is 3.06.

\input{tables/per_model_analysis}

\subsection{MTurk results}
\label{subsec:mturk_results}

We summarize the results of the crowdsourced annotations, corresponding to 2,160 approved HITs on MTurk, in \cref{tab:model_wise_results} and \cref{tab:model_error_types}. In \cref{tab:model_wise_results}, we show the ratios of ``Yes'', ``Weak yes'', ``Weak no'' and ``No'' for different models and rankings. Here, we observe that weaker results (\ie, worse ranked pairs) have lower ``Yes'' and ``Weak yes'' ratios.
For example, the annotation results for PCME show that the ``Yes'' ratios monotonically decrease as the rank goes down: 52.3\% for the most similar pairs, but 27.8\% for the least similar pairs.
Interestingly, we observe that the average ratio of ``Yes'' + ``Weak yes'' for the top-1 retrieved items exceed 80\% for all five models (\eg, 86.0\% for VSRN), while the \rone score of each model is known to less than 60\% (See Table 4 in the main paper).

From the table, we observe that by letting annotators verify more similar pairs by machines, the annotation process becomes more efficient, \ie, we can acquire the same amount of positive annotations with less number of human verification. However, as we will discuss in depth later, we emphasize that the model power is not only factor to consider: model biases emerge in MITL-produced datasets regardless of the strength of the model.

We additionally show the rationales for the uncertain matches and the specification of the errors in \cref{tab:model_error_types}. We observe that the models result in similar patterns in the annotations' rationale and specification of errors. Finally, by our annotation process, the average number of ``100\% YES'' and ``Partially YES'' images for each caption is 8.3 and 7.1, respectively. It is remarkable since original COCO annotations allow only one image to be positively paired to a caption, revealing the massive amount of missed positive matches.

\section{\ourdataset Post-processing Details}
\label{sec:eccv_postprocessing}

\subsection{The full list of invalid captions and images}
\label{subsec:invalid_items}

In this subsection, we list up the invalid captions and images in the original COCO test split. We filter the invalid captions by the following process: (1) We first list up the ``true positive'' (\ie, the positive pairs in the original COCO test set) annotated by ``100\% No'' items by Turkers or CxC \cite{parekh2020crisscrossed}. (2) We manually validate the items into two categories: totally wrong captions (\eg, ``I don't know'' captions) and semantically incorrect captions (\eg, ``A group of birds flying above the beach'' for an image with kites), The full list of invalid captions with their COCO caption ids are as follows:

{\small
\begin{itemize}
    \item[\textbullet] 607516 The first picture is blank all the time on purpose.
    \item[\textbullet] 607486 Why is my first one a blank every time.
    \item[\textbullet] 433639 There is no image here to provide a caption for.
    \item[\textbullet] 248212 I am unable to see an image above.
    \item[\textbullet] 469834 There is no image here to provide a caption for.
    \item[\textbullet] 462530 I really cant see this image very well.
    \item[\textbullet] 469102 There is no image to be reviewed on this hit.
    \item[\textbullet] 743575 There is no image showing on this page to describe.
    \item[\textbullet] 246706 I am unable to see an image above.
    \item[\textbullet] 61717 There is no picture here to describe with a caption.
    \item[\textbullet] 500797 I am unable to see the image above.
    \item[\textbullet] 19273 There is no image for me to write about.
    \item[\textbullet] 630298 There is no image to provide a caption for.
    \item[\textbullet] 576409 I am unable to see the image above.
    \item[\textbullet] 390637 I am unable to see an image above.
    \item[\textbullet] 296557 There is no image here to provide a caption for.
    \item[\textbullet] 450553 I am unable to see an image above.
    \item[\textbullet] 44809 blank image with no pictures available to write about
\end{itemize}
}

We also show the list of semantically incorrect captions as the follows:
{\small
\begin{itemize}
    \item[\textbullet] 610564 An individual is in the open view in the image.
    \item[\textbullet] 359139 I cant tell if the bears may be fighting or kissing.
    \item[\textbullet] 218995 A baseball player hugging another player as lovers do.
    \item[\textbullet] 609235 Individuals are up and doing something fun today.
    \item[\textbullet] 143250 The bar of the small bathroom has many remotes on it.
    \item[\textbullet] 375316 A photo duplicated a few times and put together.
    \item[\textbullet] 712683 Talk about a bad hair day, his is frightful.
    \item[\textbullet] 75083 A group of birds flying above the beach.
    \item[\textbullet] 625605 It is always wise to have bottles of water on hand in case of an emergency.
    \item[\textbullet] 620511 If the motorcycle brakes down, the bicycle will be good transportation.
    \item[\textbullet] 613949 A full view of an outdoor space with many things to see.
    \item[\textbullet] 129825 A picture of a comment that is open.
    \item[\textbullet] 605566 There is a room with various items in the picture.
    \item[\textbullet] 634829 That thing is really red and slow lol
\end{itemize}
}

\begin{figure}[t]
    \centering
    \includegraphics[width=0.5\linewidth]{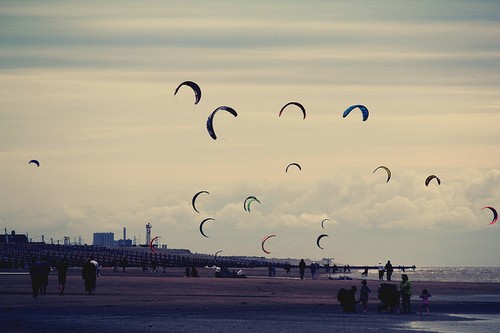}
    \caption{\small \textbf{An example image of semantically wrong captions.} We annotate ``A group of birds flying above the beach'' as a wrong caption of the figure, while the other captions are available in \cref{fig:example_eccv_more}.}
    \label{fig:wrong_caption}
\end{figure}

Finally, we omit \texttt{COCO\_val2014\_000000578492.jpg} from our test set, where the image is duplicated to training images: \texttt{COCO\_train2014\_000000388662.jpg} and \texttt{COCO\_train2014\_000000397819.jpg}.

\subsection{More examples of \ourdataset}
\label{subsec:eccv_examples}

\begin{figure}
    \centering
    \begin{subfigure}[b]{.9\linewidth}
        \centering
        \includegraphics[width=\textwidth]{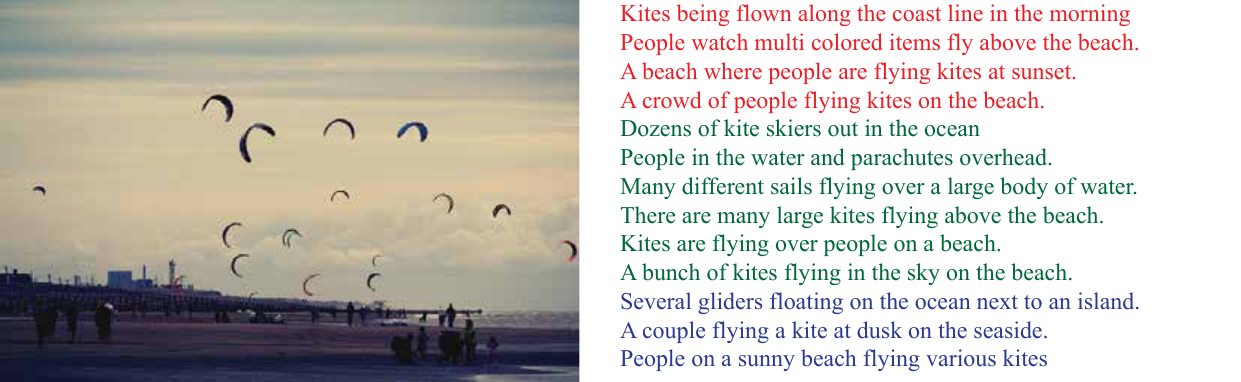}
        \vspace{.5em}
    \end{subfigure}

    \begin{subfigure}[b]{\linewidth}
        \centering
        \includegraphics[width=\textwidth]{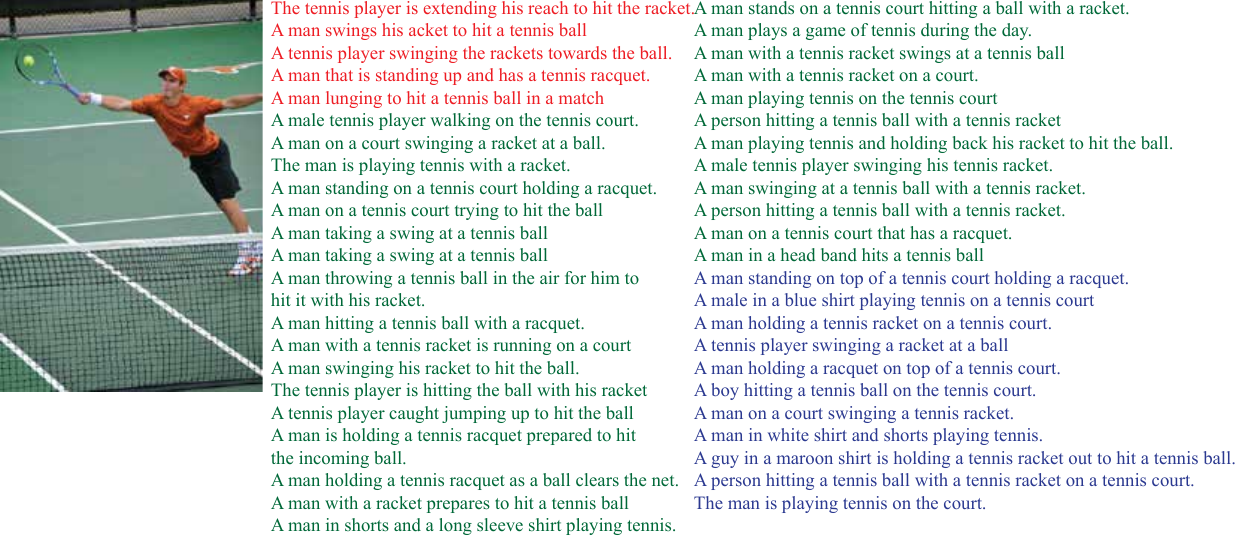}
        \vspace{.5em}
    \end{subfigure}

    \begin{subfigure}[b]{\linewidth}
        \centering
        \includegraphics[width=\textwidth]{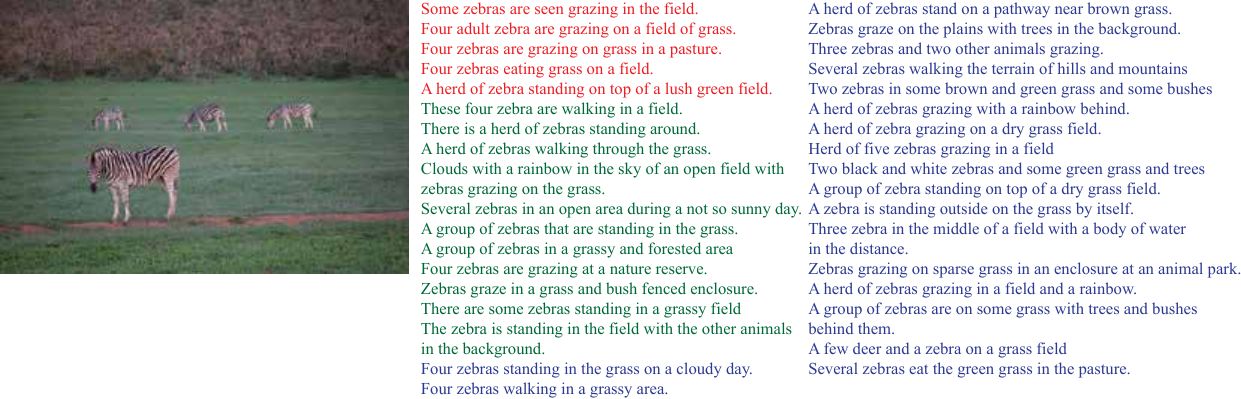}
        \vspace{.5em}
    \end{subfigure}

    \begin{subfigure}[b]{\linewidth}
        \centering
        \includegraphics[width=.8\textwidth]{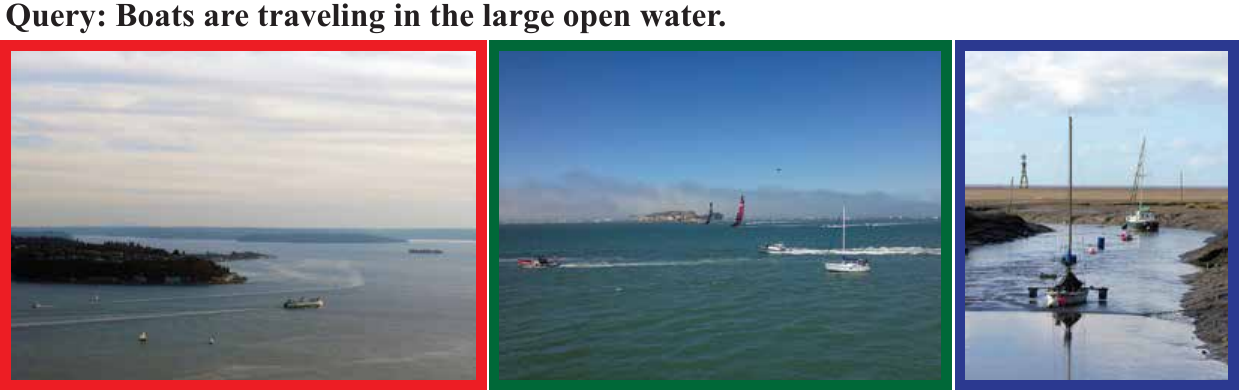}
    \end{subfigure}
    \caption{\small {\bf Example sample from \ourdataset}. Positive captions and images in \ourdataset. \textcolor{red}{Red}: original positive. \textcolor{green}{Green}: annotated as ``100\% Yes''. \textcolor{blue}{Blue}: annotated as ``Weak Yes''.}%
    \label{fig:example_eccv_more}
\end{figure}

We illustrate the samples from \ourdataset in \cref{fig:example_eccv_more}.

\section{More Evaluation Results on \ourdataset}
\label{sec:more_evaluation}

\subsection{User study for evaluation metrics}
\label{subsec:user_study_details}

\begin{figure}[ht!]
    \centering
    \begin{subfigure}[b]{\linewidth}
        \centering
        \includegraphics[width=\textwidth]{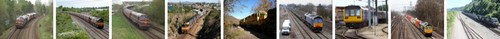}
        \caption*{\small \bf GT images for ``A train on a train track near many trees''.}
    \end{subfigure}
    \begin{subfigure}[b]{\linewidth}
        \centering
        \includegraphics[width=\textwidth]{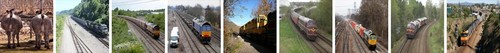}
        \caption*{\small \bf (A) Only top-1 is wrong.}
    \end{subfigure}
    \begin{subfigure}[b]{\linewidth}
        \centering
        \includegraphics[width=\textwidth]{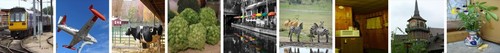}
        \caption*{\small \bf (B) Only top-1 is correct.}
    \end{subfigure}
    \begin{subfigure}[b]{\linewidth}
        \centering
        \includegraphics[width=\textwidth]{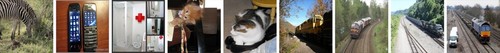}
        \caption*{\small \bf (C) Top-1 to -5 are wrong.}
    \end{subfigure}
    \begin{subfigure}[b]{\linewidth}
        \centering
        \includegraphics[width=\textwidth]{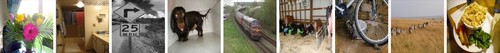}
        \caption*{\small \bf (D) Only top-5 is correct.}
    \end{subfigure}
    \begin{subfigure}[b]{\linewidth}
        \centering
        \includegraphics[width=\textwidth]{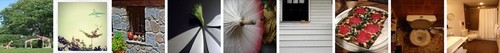}
        \caption*{\small \bf (E) All items are wrong.}
    \end{subfigure}
    \caption{\textbf{Examples of five ranking systems compared by our user study.}}
    \label{fig:user_study_ranking_examples}
\end{figure}

In this subsection, we describe the details of the user study to compare \mapr and \recallatk in terms of human judgement. We first randomly sample 40 captions from the captions whose the number of corresponding images is between 5 to 8. Then, we construct five rankings for each caption: (A) only top-1 is wrong (B) only top-1 is correct (C) top1 to 5 are wrong (D) only top-5 is correct, and (E) all items are wrong. When the number of corresponding images is 5, then we treat (C) as (E).
Each ranking system shows different \mapr and \recallatk; if we assume the number of the positives is 8, then (A) shows 0 \rone, 100 \rfive and 66.0 \mapr, (B) shows 100 \rk and 12.5 \mapr, (C) shows 0 \rk and 10.3 \mapr, and (D) shows 0 \rone, 100 \rfive and 2.5 \mapr. 
The examples of each ranking system is illustrated in \cref{fig:user_study_ranking_examples}.

\begin{figure}[ht!]
    \centering
    \includegraphics[width=\linewidth]{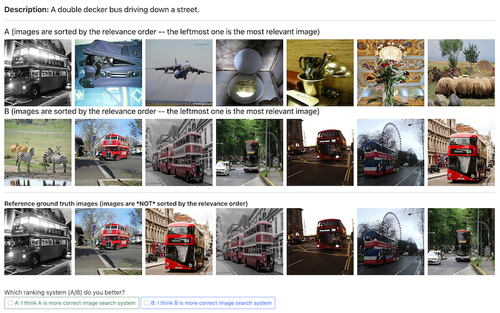}
    \caption{\small \textbf{Example question in the user study for evaluation metric comparisons.} The question asks which ranking system A or B looks more correct by humans. There are 40 of such questions in each HIT. We collect 8 participants per each question.}
    \label{fig:user_study_example}
\end{figure}

We collect binary preferences for the all possible combinations of (A) to (E), namely 10 binary pairs.
We use MTurk for collecting participants, and we collect 8 participants for each question (the example question is in \cref{fig:user_study_example}). As a result, we collect $40 \times 10 \times 8 = 3,200$ binary preferences of the five different rankings. We list the full binary preference in \cref{tab:full_binary_preference}.
After collecting binary preferences, we restore the preference rankings using Bradley–Terry (BT) model \cite{bradley1952rank}. The BT model assumes that for the given pair $i$ and $j$, the probability to the pairwise comparison $i > j$ is linear to the true ranking score, \ie, $P(i > j) = \frac{p_i}{p_i + p_j}$. Our goal is to estimate $p_i$, the true ranking preference for each method. Using BT model, we got the following results: A (70.85), B (13.15), C (10.66), D (4.89). This confirms that \mapr is more aligned to humans than \rone; (A) shows 0 \rone and B shows 100 \rone, however humans prefer (A) to (B) where (A) has higher \mapr than (B) (66.0 and 12.5 if the number of positives is 8).

\begin{table}[ht!]
    \centering
    \small
    \setlength{\tabcolsep}{6pt}
    \begin{tabular}{@{}lccccc@{}}
    \toprule
      & A   & B   & C   & D   & E   \\ \midrule
    A: only top-1 is wrong  & 0   & 231 & 66  & 111 & 316 \\
    B: only top-1 is correct & 89  & 0   & 21  & 79  & 328 \\
    C: top-1 to -5 are wrong & 254 & 299 & 0   & 273 & 343 \\
    D: only top-5 is correct & 185 & 217 & 23  & 0   & 287 \\
    E: all items are wrong & 28  & 16  & 1   & 9   & 24  \\ \bottomrule
    \end{tabular}
    \vspace{.5em}
    \caption{\small \textbf{Binary preferences for five ranking systems.} Each number in row $i$ and column $j$ denotes that the number of preferences $i > j$. For example, 231 responses preferred ``(A) only top-1 is wrong'' than ``(B) only top-1 is correct'', while the number of the converse case is 89.}
    \label{tab:full_binary_preference}
\end{table}

\subsection{Training details}
\label{subsec:training_details}

We follow the implementation details by Chun \etal \cite{chun2021pcme}. We use the AdamP \cite{heo2021adamp} optimizer and cosine annealing learning rate scheduling \cite{loshchilov2016sgdr}.
For re-implemented VSE, PVSE and PCME, we use the pre-trained ResNet-152 backbone and pre-trained Glove vectors following previous studies \cite{faghri2018vsepp, song2019pvse, chun2021pcme}. We use two-stage training scheme that includes ``pre-training'' (freezing pre-trained backbones, but only updating the additional modules) and ``fine-tuning'' (updating the whole parameters). The models are pre-trained during 30 epochs and fine-tuned during 30 epochs. For the improved PCME, we use the ResNet-152 model trained with the CutMix \cite{cutmix} augmentation for achieving better \rone accuracies.

\subsection{Full table}
\label{subsec:more_evaluation_full_table}

\input{tables/main_results_detail}

\cref{tab:main_results_it} and \cref{tab:main_results_ti} show the full results of each model for image-to-text retrieval tasks and text-to-image retrieval tasks, respectively.

\section{Analysis of Biases in MITL}
\label{sec:bias_analysis}

In this section, We explore and quantify the effect of the choice of multiple machine annotators to the dataset quality.
We delve into the effect of the choice of machine annotators used for machine-in-the-loop (MITL) labeling paradigm to the dataset quality. Specifically, we are interested in the model bias, the type of bias that arises because of the pre-selection of plausible samples by the model in the annotation pipeline.
We discuss the generalizability of our framework to general annotation tasks in \cref{subsec:annotation_tasks}, and the definition of ``bias'' in our dataset process in \cref{subsec:what_is_bias}.
We measure the model bias by employing the crowdsourced data as the source for evaluating the ITM models. For a perfectly unbiased data, we shall expect the identical rankings across the version of datasets collected with different models.
We show the performances on different versions of the datasets using only one MITL model and provide discussions (\cref{subsec:biases_in_eccv_caption}).

\subsection{Image caption matching problem to general annotation tasks}
\label{subsec:annotation_tasks}

Many real-world applications are powered by state-of-the-art machine learning (ML) models shown to exceed human-level performances in tasks such as natural language understanding \cite{bert} and image classification \cite{resnet,geirhos2018generalisation}. However, previous studies have shown that two conditions must be met for these models to perform well: massive training data and quality annotations.
For example, large datasets that consist of well-curated 1M images \cite{imagenet}, 3.5B Instagram photos \cite{mahajan2018exploring, singh2022revisiting}, 300M web photos \cite{vit}, 400M captioned images \cite{clip}, 1.8B noisy captioned images \cite{jia2021scaling}, 15M hierarchically structured images \cite{yun2021relabel, vit}, and synthetic images \cite{mixup, cutmix, cubuk2019autoaugment, cubuk2020randaugment} are the key factors behind the corresponding models' success.
Moreover, the annotation quality is equally important for the model performances. 
Mahajan \etal \cite{mahajan2018exploring} showed that training on 940M images with well-processed 1.5k labels results in a model comparable to training on 3.5B images with noisy and weak 17k labels; both models show 84.2\% ImageNet-1K top-1 accuracy.

An emerging pattern for obtaining quality labels for a large dataset is pipelining a machine learning model and human annotators. Expert human annotators are reliable and produce high-quality label, but they are costly to accommodate. Strong machine annotators are relatively inexpensive, but they result in low-quality and unreliable annotations. 
One popular method of combining the two is feeding human annotators with machine learning model's outputs. For example, a model suggests annotations (\eg, candidates of labels \cite{openimagesv4}, estimated boxes \cite{openimagesv4}, estimated segmentation maps \cite{OpenImagesV5}, estimated descriptions \cite{kayser2021evil}) for the given data point, and the annotators only need to confirm or fix the labels given by the machine annotators. This approach is commonly used for building a large-scale dataset, such as OpenImages \cite{openimagesv4, OpenImagesV5}, e-Vil \cite{kayser2021evil}.

While rich body of discussions are available for building annotation interfaces and crowdsourcing workflows, we still lack a good understanding of the impact the underlying machine learning models have on the annotators and on the annotation results. In this research, we specifically examine the downstream effects of a common practice where researchers and practitioners consider only one ``strong'' model in the machine-in-the-loop annotation pipeline.
This can be problematic because different models not only show different results to the human annotators, but also in different orders, bring the impartialness of the annotated dataset towards any particular model to the surface.
In other words, the machine-in-the-loop annotations are not stable across model choices.

As a realistic scenario for utilizing ML models to aid annotators, we consider the COCO Caption matching task \cite{lin2014microsoft, chen2015microsoft} that matches each image with sentences in a large database of captions. Due to the sheer bulk of the involved databases (123,287 images and 616,767 captions), it is infeasible for annotators to search through the matching caption. Instead, for each image, we use model-based ranking of possible captions to greatly reduce the search space for annotators. We have conducted studies with five state-of-the-art image-text matching models.
Our COCO Caption matching task can be seen as the extreme version of the class label selection task where the number of possible classes is as large as the number of possible descriptions.

A decent overview of the image annotation tools is provided by Sager \etal \cite{sager2021survey}. The annotation task is determined along two axes: the types of inputs and the expressiveness of the labels. This paper focuses on the image inputs, one of the most frequently annotated type of data. The expressiveness and complexity of labels are directly related to the learning task being addressed. Tagging images with class labels is arguably the most common and basic form in the spectrum of label expressiveness. In the other extreme, we have the \textit{image-caption matching task}: given an image, annotator has to search through the database of descriptions to find the one that best matches the image \cite{chen2015microsoft}. The caption matching task is of the same nature as image tagging: one needs to find the correct label in a list of possible labels. However, the candidate space for the possible labels is exponentially greater for the caption matching task. If the vocabulary size is $V$ and the lengths of caption sequences are generally $L$, the size of the candidate space is as large as $O(V^L)$. This contrasts against the number of possible class labels that are generally far smaller than $V$. 

We consider the image-caption matching as a testbed for analyzing annotation pipelines for two reasons.  First, it highly relevant for the MITL annotation paradigm because it is downright infeasible for humans to browse through the database. Second, it inherits the same tool as image tagging, making our experimental results and analyses transferable to general image tagging tasks. 

\subsection{What Do We Mean by ``Bias''?}
\label{subsec:what_is_bias}

Bias is an overloaded term with multiple senses. We briefly explore its use in relevant fields and make a definition relevant to our paper.

In statistics and machine learning, bias of an estimator or model refers to the mismatch between its average behavior and the true parameter or underlying function \cite{johnson2000probability,bishop2006pattern}. We partially adopt this definition of bias in a broad sense. The annotation pipeline as a whole can be regarded as a mechanism for assigning plausible labels to a given set of images. When we say that the annotation pipeline is ``biased'', we refer to the discrepancy between the resulting annotations and the true, underlying labels for the samples.

In human-related studies, like psychology, neuroscience, human-computer interaction, and increasingly in machine learning, the use of ``bias'' often points to its underlying human factor. Examples include ``confirmation bias'' where humans favorably select data that serve their purpose \cite{plous1993psychology}, ``reporting bias'' where crucial commonsense knowledge is overlooked \cite{easterbrook1991publication}, and ``survivorship bias'' where non-surviving cases are under-represented \cite{mangel1984abraham}. In our MITL annotation pipeline, we study the \textit{model bias} where models hinder humans from generating an unbiased set of labels by presenting humans with only a selection of the candidate labels deemed correct by the models.

\subsection{Biases in \ourdataset}
\label{subsec:biases_in_eccv_caption}

Given the crowdsourced labeled image-caption data of \ourdataset, our aim is to analyze the degrees of bias in them, depending on the underlying model used for machine-in-the-loop (MITL) labeling paradigm. Specifically, we are interested in the model bias, the type of bias that arises because of the pre-selection of plausible samples by the model in the annotation pipeline. We measure the model bias by employing the crowdsourced data as the source for evaluating the cross-modal retrieval models. For a perfectly unbiased data, we shall expect the identical evaluation results (\ie, in terms of the ranking of the methods) across the version of datasets collected with different models. In this section, we introduce the strategy to measure the model bias and present experimental results and analyses.

We measure the model bias in a labeled dataset by examining whether certain versions of datasets behaves favorably to certain models when they are used as the evaluation benchmark. To measure this, we need to introduce the specific evaluation metrics used for measuring the cross-modal retrieval performances. We use \textit{Recall@1} and \textit{R-Precision}. 

Recall@1 is the most widely-used metric for reporting the performances of cross-modal retrieval models.
To compute it, we first let $m_i (x)$ be the indicator whether the $i$-th retrieved item of the input $x$ is a positive match. 
\begin{align}
\label{eq:matching}
\begin{split}
m_i (x) &= 1 \quad \text{if $i$-th retrieved item of $x$ is a positive match;}\\
m_i (x) &= 0 \quad \text{otherwise.}
\end{split}
\end{align}
\rone measures whether the top-1 retrieved item is a positive match on average:
\begin{equation}
\label{eq:r1}
    \text{R@1} = \frac{1}{N} \sum_{n=1}^N m_1 (x_n).
\end{equation}
Despite its popularity, Recall@1 has a serious shortcoming. As argued by Musgrave \etal \cite{musgrave2020metric}, a high Recall@1 does not always guarantee high-quality retrieval results. Musgrave \etal have proposed to use the R-Precision as an alternative metric. Let $R(x)$ be the total number of matched items for the input $x$. Then, R-Precision is defined as follows: 
\begin{equation}
\label{eq:rp}
    \text{R-P} = \frac{1}{N}\sum_{n=1}^N \frac{1}{R(x)} \sum_{i=1}^{R(x)} m_i(x_n).
\end{equation}
Despite its good properties, in practice, it is impossible to use R-Precision for cross-modal retrieval benchmarks because many cross-modal retrieval benchmarks only a few number of (\eg, 1) positive pairs in the dataset. However, as shown in \cref{fig:coco_retrieval_results}, there actually are many plausible positive pairs missed by the original positive pairs.

We further refurbish the metrics by using the fine-grained degrees of pair positivity provided by the annotators (\cref{subsec:mturk_annotators}), which are not available for conventional cross-modal retrieval datasets.
We update the matching function (\cref{eq:matching}) as follows:
\begin{align}
\label{eq:matching_mturk}
\begin{split}
m_i (x) &= 1 \quad \text{if $i$-th retrieved item of $x$ is annotated as ``100\% YES''}\\
m_i (x) &= 0.5 \quad \text{if $i$-th retrieved item of $x$ is annotated as ``Partially YES''}\\
m_i (x) &= 0 \quad \text{otherwise}.
\end{split}
\end{align}
We report the modified Recall@1 and R-Precision using the matching function above as the final performance metric for cross-modal retrieval models on our crowdsourced datasets.

\input{tables/model_performances_vs_machine_annotators}

\cref{tab:main_r1} and \cref{tab:main_pr} show the performances on different versions of the datasets using only one MITL model.
Not surprisingly, we observe that the best-performing model for each dataset coincides with the model used for the MITL label proposal (\ie, the diagonal elements in Table \ref{tab:main}). Other models tend to show quite some drop in performance. This strongly corroborates the existence of model bias in datasets collected with the aid of machine filtering. We further observe that even for models not used for generating the label proposals (\ie, the non-diagonal elements in \cref{tab:main_r1} and \cref{tab:main_pr}), the rankings shift with respect to the underlying MITL model. This suggests that even when one avoids the direct use of the MITL model for evaluating its own performance, one may still observe unstable evaluation results, where different MITL models arbitrarily favor different models.

\begin{figure}[t!]
    \centering
    \includegraphics[width=\linewidth]{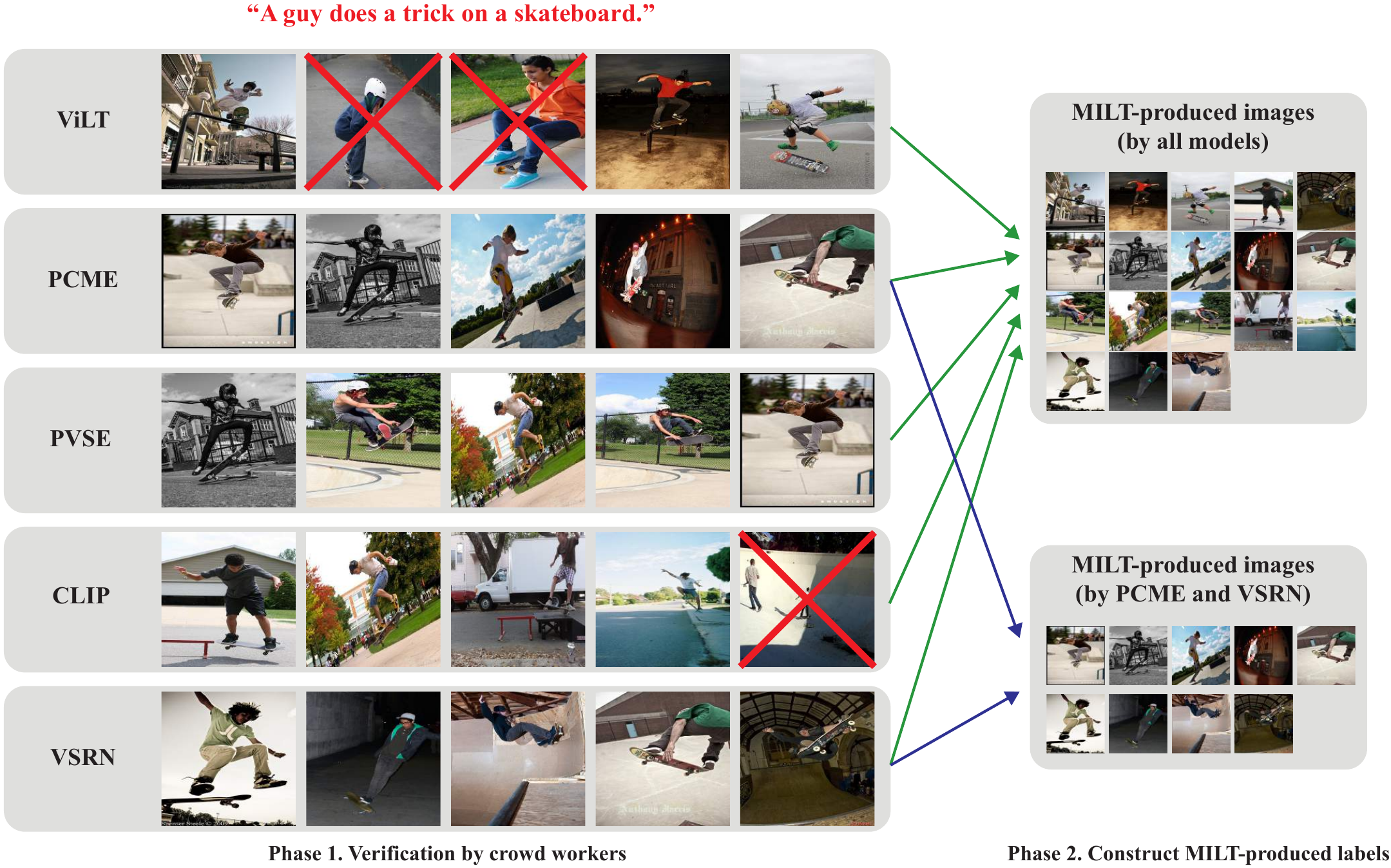}
    \caption{\textbf{Overview of our machine-in-the-loop annotation process.} We choose the subset of the verified image caption pairs by crowd workers to control the effect by the models to final annotations.}
    \label{fig:ann_process}
\end{figure}

This suggests that even when one avoids the direct use of the MITL model for evaluating its own performance (\ie, avoiding the diagonal results), one may still observe unstable evaluation results, where different MITL models arbitrarily favor different models.

\begin{figure}[t!]
    \centering
    \includegraphics[width=.5\linewidth]{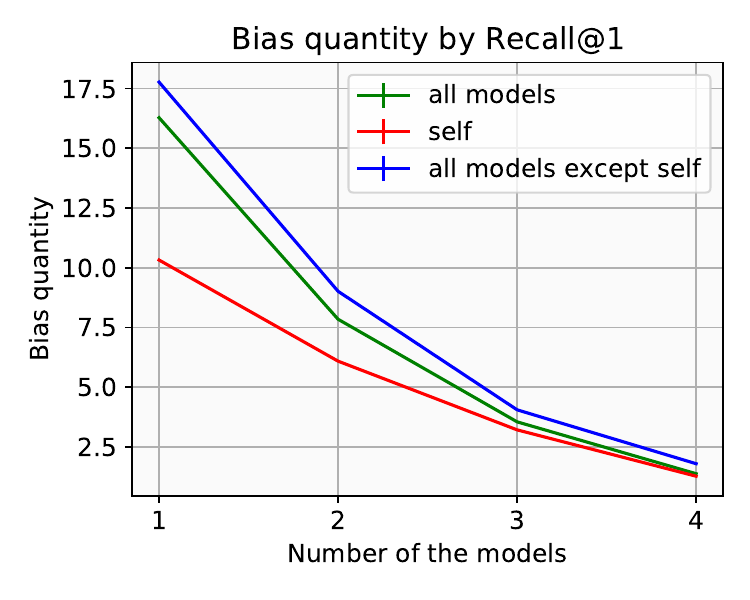}%
    \includegraphics[width=.5\linewidth]{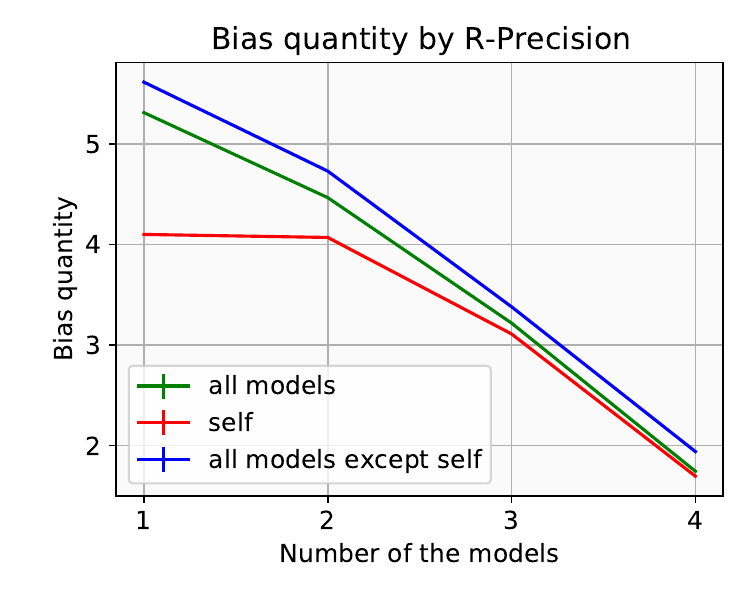}
    \caption{\textbf{Number of models vs. bias quantity.} The bias quantities (Eq. \eqref{eq:bias_quantify}) with changing the number of models for the annotation filtering process are shown. For both Recall@1 and R-Precision metrics, we observe that using more models reduce the severity of bias; the discrepancy between the resulting annotations by models and the underlying labels for the samples decreases. 
    }
    \label{fig:bias_vs_num_models_appendix}
\end{figure}

A crucial limitation in an analysis of this type is the lack of the \textit{true labels}. The obtained versions of datasets are clearly enhanced versions compared to the original COCO Caption dataset, but they are still heavily affected by the model biases as seen above. To make a better estimate of how far the datasets are from the true labels, we introduce the \textit{multi-model strategy} where the workers verify label proposals generated by more than two of the models involved. More specifically, a multi-model strategy involving models $\Theta=\{\text{PVSE},\text{PCME}\}$ pools the label proposals from both PVSE and PCME and present them to the human annotators. The rest of the verification process is identical as before.
The intuition is that the dataset built with the multiple models will be much closer to the true labels for the image-caption matches. In the extreme case, we have the \textit{all-model strategy} involving all five models considered in this work. See the ``All'' columns in \cref{tab:main_r1} and \cref{tab:main_pr} for the corresponding results.

Based on this intuition, we additionally measure the distance between a version of dataset and the all-model dataset, which is deemed to contain the labels that are closer to the true labels. We define $s_\Theta(\phi)$ as the performance of the model $\phi$ evaluated upon the dataset built with the multi-model strategy with MITL models $\Theta$. We write $s_\text{All}(\phi)$ as a good proxy for the true performance of the model $\phi$. We define the model bias incurred by a subset of models $\Theta$ as 
\begin{align}
    \mathcal B_\Theta := \frac{1}{5}\sum_{\phi\in\text{All}} \left|s_\Theta(\phi) - s_\text{All}(\phi)\right|
    \label{eq:bias_quantify}
\end{align}
where $\text{All}= \{ \text{PVSE, VSRN, PCEM, ViLT, CLIP} \}$.
For example, 
$ \mathcal B_{\{\text{PVSE}\}}$ using Recall@1 (\cref{tab:main_r1}) is computed as follows: 
\begin{align}
\label{eq:bias_function_for_list}
\begin{split}
    \mathcal B_{\{\text{PVSE}\}}&=(| 76.5 - 76.6 | + | 68.0 - 80.1 | + | 67.7 - 77.4 |\\
    &+ | 59.5 - 72.4 | + | 51.7 - 64.3 |) / 5 = 9.5
\end{split}
\end{align}
We break down the degree of bias $\mathcal B_\Theta$ into the bias incurred onto oneself (``self-bias'') and the one incurred onto the other models (``non-self-bias''). We quantify the self-bias for a set of models $\Theta$ to measure the amount of the bias incurred onto oneself: $\frac{1}{|\Theta|}\sum_{\phi\in\Theta} \left|s_\Theta(\phi) - s_\text{All}(\phi)\right|$. For example, self-bias for PVSE is $| 76.5 - 76.6 | = 0.1$. The complementary amount of bias, the non-self-bias is computed similarly. For example, PVSE's non-self-bias is computed as $| 68.0 - 80.1 | + | 67.7 - 77.4 | + | 59.5 - 72.4 | + | 51.7 - 64.3 |) / 4 = 11.8$.

We plot the degrees of the biases, measured with $\mathcal B_\Theta$,  self-bias, and non-self-bias in \cref{fig:bias_vs_num_models_appendix}. We have experimented with changing the numbers of models $|\Theta|$. All numbers are averaged over all possible subsets of models of size $|\Theta|$ (\eg, if $| \Theta | = 2 $, the result is the average of $n(n-1)/2$ numbers). We omit the case with $|\Theta|=5$ because all metrics are zero by definition.

\begin{figure}[t!]
    \centering
    \includegraphics[width=\textwidth]{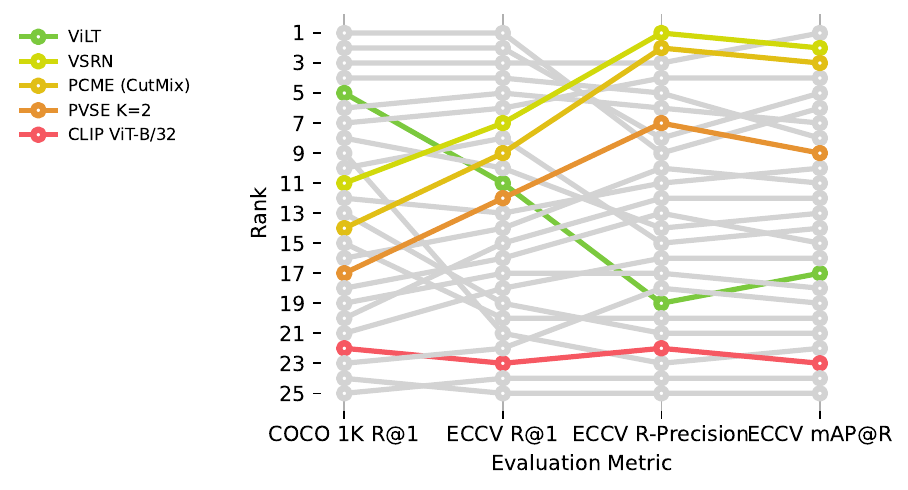}
    \caption{\textbf{Bias quantity in the dataset.} 
    The full rankings of the chosen MITL annotators on our benchmark.}
    \label{fig:bias_quantity_ranking}
\end{figure}

We have two observations. First, the smaller number of MITL models make the gap between ``self-bias'' and ``model-bias'' larger. This implies that if we use a single model for the MITL annotations, the resulting dataset is highly unlikely to treat the methods differently. 
Second, the general bias measurements decrease with the number of involved models in the MITL annotation process. For the practical applications, hence, it is advisable to use multiple models to collect the label candidates to verify.
In practice, we observe that only three machine annotators (PVSE, PCME and VSRN) achieve better rankings on \eccv \mapr compared to the COCO \rone ranking (\cref{fig:bias_quantity_ranking}).
That suggests that our \ourdataset is not fully biased towards the selected machine annotators.

\section{Noisy Annotations}
\label{sec:noisy_annotations}

\begin{figure}[t!]
    \centering
    \begin{subfigure}[b]{\linewidth}
        \centering
        \includegraphics[width=\textwidth]{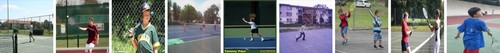}
        \caption{\small \bf ``A boy holding a tennis racquet on a tennis court.''.}
        \label{fig:failure_examples_a}
    \end{subfigure}
    \begin{subfigure}[b]{\linewidth}
        \centering
        \includegraphics[width=\textwidth]{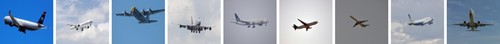}
        \caption{\small \bf ``A large white airplane flies in the gray sky.''.}
        \label{fig:failure_examples_b}
    \end{subfigure}
    \begin{subfigure}[b]{\linewidth}
        \centering
        \includegraphics[width=\textwidth]{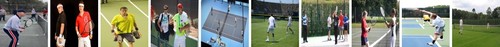}
        \caption{\small \bf ``Two men of some sort on a tennis court.''.}
        \label{fig:failure_examples_c}
    \end{subfigure}
    \caption{\small \textbf{Examples of noisy annotations in \ourdataset.} Examples of false positive images are shown. Each of false positive contains (a) wrong object, \eg, ``baseball bat'' instead of ``tennis racquet'' (b) wrong color, \eg, ``blue'' instead of ``gray'' (c) wrong quantity, \eg, ``one'' instead of ``two''.}
    \label{fig:failure_examples}
\end{figure}

Our annotations are built upon crowdsoure annotations. Due to the nature of the noisiness of crowdsource annotations, \ourdataset contains some wrong annotations. Also, our annotations are chosen not only from ``100\% YES'' but also from ``Partially YES''. Note that our HIT is designed for specifying the details of what makes the annotation ``partially correct'' -- See \cref{fig:mturk_example}.
We illustrate some false positive cases in \cref{fig:failure_examples}. The false positives can be happened due to (1) wrong object, \eg, ``baseball bat'' instead of ``tennis racquet'' (\cref{fig:failure_examples_a}), (2) wrong color, \eg, ``blue'' instead of ``gray'' (\cref{fig:failure_examples_b}) (c) wrong quantity, \eg, ``one'' instead of ``two'' (\cref{fig:failure_examples_c}).
However, although there exist some false positives in our dataset, we strongly encourage to use \ourdataset and \mapr for evaluating a new VL model. Even if there exist some false positives, they are not 100\% wrong examples; if a model learns good global ranking, then the partially correct examples should be located in the top rankings than other random items. Therefore we strongly encourage to use \mapr instead of \recallatk; \mapr can mitigate the error by false positives, while \recallatk can amplify errors by noisy annotations by only checking whether the top-K retrieved items are in the true items or not.

%% file: tables/model_kendall.tex
\begin{table}[t]
\begin{subtable}[t]{0.48\linewidth}
\setlength{\tabcolsep}{1pt}
\small
\centering
\begin{tabular}[t]{@{}cccccc@{}}
\toprule
     & PVSE  & VSRN  & PCME  & ViLT   & CLIP   \\ \midrule
PVSE & -     & 0.266 & 0.430 & 0.089  & 0.088  \\
VSRN & 0.266 & -     & 0.260 & 0.060  & 0.049  \\
PCME & 0.430 & 0.260 & -     & 0.110  & 0.120  \\
ViLT & 0.089 & 0.060 & 0.110 & -      & -0.013 \\
CLIP & 0.088 & 0.049 & 0.120 & -0.013 & -      \\ \bottomrule
\end{tabular}
\caption{\small {\bf Model similarity analysis by Kendall's $\tau$.} A higher score means that two models are more correlated.}
\label{tab:model_kendall}
\end{subtable}
\hspace{\fill}
\begin{subtable}[t]{0.48\linewidth}
\small
\centering
\begin{tabular}[t]{@{}lccccc@{}}
\toprule
     & PVSE  & VSRN & PCME & ViLT & CLIP \\ \midrule
PVSE & -     & 4.52 & 3.87 & 5.95 & 6.24 \\
VSRN & 4.50  & -    & 4.47 & 5.45 & 5.97 \\
PCME & 3.87  & 4.50 & -    & 5.79 & 6.20 \\
ViLT & 5.79  & 5.19 & 5.78 & -    & 6.18 \\
CLIP & 6.16  & 5.81 & 6.12 & 6.03 & -  \\ \bottomrule
\end{tabular}
\caption{\small {\bf Model similarity analysis by the average ranking.} A smaller rank means that two models are more similar.}
\label{tab:model_avg_rank}
\end{subtable}
\caption{\small {\bf Model similarity analyses.} We measure similarties between the machine annotators in two different ways. (a) We measure the ranking correlations using Kendall's $\tau$; 1.0 means two lists are identical, -1.0 indicates two lists are strongly disagreed each other. (b) We measure the average rankings of the image retrieved by a model for other models. Each row indicates the average ranking of the top-1 retrieved image of the row model for other column models.}
\label{tab:model_diff}
\vspace{-1.5em}
\end{table}

%% file: tables/worker_stats.tex
\begin{table}[t]
\small
\centering
\setlength{\tabcolsep}{6pt}
\begin{tabular}{@{}ccccc@{}}
\toprule
\# HITs / worker & \# Workers & \# Submit. HITs & \# Approv. HITs & Approve ratio \\ \midrule
1 & 599 & 599 & 459 & 76.6\% \\
1 $<$ and $\leq$ 5  & 254 & 681 & 535 & 78.6\% \\
5 $<$ and $\leq$ 10 & 61 & 475 & 374 & 78.7\% \\
10 $<$ and $\leq$ 15 & 23 & 297 & 216 & 72.7\% \\
15 $<$ and $\leq$ 25 & 21 & 411 & 285 & 69.3\% \\
25 $<$ & 12 & 506 & 293 & 57.9\% \\ \bottomrule
\end{tabular}
\vspace{.5em}
\caption{\small {\bf MTurk worker statistics.} The number of the unique workers, the submitted HITs, the approved HITs and the average approve ratio by the number of completed HITs are shown.}
\label{tab:worker_stats}
\end{table}

%% file: tables/per_model_analysis.tex
\begin{table}[t!]
\centering
\small
\setlength{\tabcolsep}{4pt}
\begin{tabular}{@{}llccccccccc@{}}
\toprule
        && \multicolumn{4}{c}{PVSE}       & & \multicolumn{4}{c}{VSRN}        \\
Rank && Yes  & Weak yes & Weak no & No  && Yes  & Weak yes & Weak no & No  \\  \midrule
1 && 52.3 & 31.7 & 11.2 & 4.7 && 54.4 & 31.6 & 10.0 & 4.0 \\
2 && 39.0 & 36.7 & 16.7 & 7.5 && 39.9 & 37.6 & 15.8 & 6.7 \\
3 && 34.5 & 36.9 & 18.0 & 10.6 && 36.6 & 36.8 & 17.6 & 9.0 \\
4 && 30.9 & 39.2 & 19.6 & 10.4 && 33.2 & 38.1 & 17.6 & 11.1 \\
5 && 28.0 & 39.2 & 20.5 & 12.3 && 30.4 & 38.6 & 19.0 & 12.0 \\ \midrule
Avg. && 36.8 & 36.8 & 17.3 & 9.1 && 38.8 & 36.6 & 16.0 & 8.6 \\ \bottomrule
\\
\end{tabular}
\begin{tabular}{@{}llccccccccc@{}}
\toprule
        && \multicolumn{4}{c}{PCME}       & & \multicolumn{4}{c}{ViLT}        \\
Rank && Yes  & Weak yes & Weak no & No  && Yes  & Weak yes & Weak no & No  \\  \midrule
1 && 52.3 & 33.2 & 10.4 & 4.2 && 47.6 & 33.7 & 12.0 & 6.7 \\
2 && 39.8 & 37.2 & 15.1 & 7.9 && 36.6 & 33.5 & 19.7 & 10.2 \\
3 && 35.6 & 39.0 & 17.5 & 7.9 && 30.3 & 35.6 & 20.2 & 13.9 \\
4 && 31.9 & 37.8 & 21.1 & 9.2 && 26.7 & 35.6 & 22.8 & 14.9 \\
5 && 27.8 & 40.0 & 21.5 & 10.6 && 25.8 & 35.6 & 22.5 & 16.1 \\ \midrule
Avg. && 37.4 & 37.5 & 17.2 & 8.0 && 32.9 & 34.8 & 19.7 & 12.5 \\ \bottomrule
\\
\end{tabular}
\begin{tabular}{@{}llcccc@{}}
\toprule
        && \multicolumn{4}{c}{CLIP}\\
Rank && Yes  & Weak yes & Weak no & No  \\  \midrule
1 && 50.5 & 33.5 & 12.0 & 3.9 \\
2 && 35.3 & 38.5 & 18.0 & 8.2 \\
3 && 33.8 & 39.9 & 16.1 & 10.1 \\
4 && 31.0 & 38.6 & 18.0 & 12.5 \\
5 && 28.4 & 39.8 & 20.2 & 11.6 \\ \midrule
Avg. && 36.0 & 38.0 & 16.8 & 9.2 \\ \bottomrule
\end{tabular}
\vspace{.5em}
\caption{\textbf{Model-wise annotation overview.} The percentages of ``100\% YES'', ``Partially YES'', ``Mostly NO'' and ``100\% NO'' for each model and each ranking are shown. For example, the first row indicates the annotation results for the top-1 retrieved image and description pairs.}
\label{tab:model_wise_results}
\end{table}

\begin{table}[t]
\small
\setlength{\tabcolsep}{6pt}
\begin{subtable}[t]{\textwidth}
\centering
\begin{tabular}{@{}lcccc@{}}
\toprule
     & Not in image & Not in caption & Incorrect object description & grammar error \\ \midrule
PVSE & 30.4         & 17.5           & 48.0                  & 4.1           \\
VSRN & 30.1         & 18.4           & 46.5                  & 5.0           \\
PCME & 31.6         & 17.9           & 46.1                  & 4.4           \\
ViLT & 31.2         & 18.9           & 45.7                  & 4.2           \\
CLIP & 28.8         & 18.5           & 47.1                  & 5.6           \\ \midrule
Avg. & 30.4         & 18.2           & 46.7                  & 4.7           \\ \bottomrule
\end{tabular}
\caption{\small \textbf{Shortcomings by models.}}
\end{subtable}
\begin{subtable}[t]{\textwidth}
\centering
\resizebox{\columnwidth}{!} {
\begin{tabular}{@{}lcccccccccc@{}}
\toprule
     & Quantity & Color & Species & Gender & Age & Action & Emotion & Location & Orientation & Others \\ \midrule
PVSE & 25.7     & 20.1  & 11.9    & 5.4    & 2.2 & 14.9   & 0.9     & 3.4      & 12.1        & 3.2    \\
VSRN & 23.8     & 20.1  & 12.7    & 4.0    & 1.7 & 15.5   & 1.2     & 4.3      & 13.6        & 3.0    \\
PCME & 24.8     & 20.9  & 12.1    & 5.6    & 2.1 & 14.3   & 0.8     & 4.2      & 12.1        & 3.1    \\
ViLT & 24.9     & 20.8  & 11.2    & 6.3    & 2.1 & 14.9   & 0.9     & 3.7      & 12.8        & 2.4    \\
CLIP & 27.2     & 19.7  & 13.5    & 3.5    & 1.7 & 14.5   & 1.1     & 3.9      & 11.3        & 3.5    \\ \midrule
Avg. & 25.3     & 20.3  & 12.3    & 5.0    & 2.0 & 14.8   & 1.0     & 3.9      & 12.4        & 3.0    \\ \bottomrule
\end{tabular}
}
\caption{\small \textbf{Detailed errors by models.}}
\end{subtable}
\caption{\textbf{Error types.} The percentage of the error types by models. There is no statistical significant difference by models.}
\label{tab:model_error_types}
\end{table}

%% file: tables/main_results_detail.tex
\begin{table}[t!]
\small
\centering
\setlength{\tabcolsep}{4pt}
\resizebox{\columnwidth}{!} {
\begin{tabular}{@{}lcccccccccccccc@{}}
\toprule
 & \multicolumn{3}{c}{ECCV} &  & CxC &  & \multicolumn{3}{c}{COCO 1K} &  & \multicolumn{4}{c}{COCO 5K} \\
 & mAP@R & R-P & R@1 &  & R@1 &  & R@1 & R@5 & R@10 &  & R@1 & R@5 & R@10 & PMRP \\  \midrule
\multicolumn{15}{c}{ResNet-152 \cite{resnet} image encoder + Bi-GRU \cite{cho2014properties} text encoder} \\ \midrule
VSE0$^\dagger$ & 14.92 & 26.05 & 44.73 &  & 26.92 &  & 48.50 & 81.74 & 89.26 &  & 24.76 & 53.82 & 67.98 & 51.54 \\
VSE++$^\dagger$ & 24.45 & 36.51 & 64.31 &  & 43.66 &  & 66.86 & 91.34 & 95.80 &  & 41.52 & 72.10 & 82.88 & 59.58 \\
PVSE K=1 & 23.40 & 35.56 & 62.57 &  & 43.88 &  & 66.70 & 90.94 & 95.60 &  & 41.76 & 72.96 & 82.90 & 58.80 \\
PVSE K=2 & 26.72 & 39.24 & 65.03 &  & 46.08 &  & 68.42 & 91.26 & 96.18 &  & 44.10 & 73.38 & 83.68 & 60.69 \\
PCME & 26.24 & 38.65 & 65.50 &  & 46.22 &  & 68.78 & 91.42 & 96.38 &  & 44.26 & 73.52 & 83.44 & 60.87 \\
PCME (CutMix)$^\dagger$ & 28.55 & 41.23 & 68.75 &  & 47.14 &  & 68.72 & 92.58 & 96.36 &  & 45.04 & 75.12 & 84.66 & 61.85 \\ \midrule
\multicolumn{15}{c}{Region features based on Bottom-up Attention \cite{Anderson2017up-down} and SCAN \cite{lee2018scan}} \\ \midrule
VSRN & 30.77 & 42.89 & 73.83 &  & 55.06 &  & 76.20 & 94.76 & 97.86 &  & 53.02 & 81.12 & 89.42 & 62.41 \\
VSRN + AOQ & 30.70 & 42.61 & 73.12 &  & 56.86 &  & 77.50 & 95.44 & 98.38 &  & 55.14 & 83.30 & 90.80 & 62.47 \\
CVSE & 28.11 & 40.14 & 69.23 &  & 52.92 &  & 74.14 & 94.94 & 98.04 &  & 51.00 & 79.58 & 89.36 & 62.23 \\
SGR & 27.24 & 39.15 & 71.05 &  & 58.80 &  & 77.26 & 95.94 & 98.26 &  & 57.24 & 83.18 & 90.64 & 61.98 \\
SAF & 27.36 & 39.30 & 71.05 &  & 56.98 &  & 78.06 & 95.84 & 98.20 &  & 55.48 & 83.82 & 91.82 & 62.00 \\
VSE infty (BUTD region) & 31.36 & 42.78 & 74.86 &  & 60.12 &  & 79.64 & 96.38 & 98.60 &  & 58.34 & 85.32 & 92.34 & 62.98 \\
VSE infty (BUTD grid) & 31.68 & 43.08 & 76.53 &  & 60.64 &  & 80.42 & 96.78 & 98.86 &  & 59.10 & 85.90 & 92.82 & 62.79 \\
VSE infty (WSL grid) & 34.80 & 45.41 & 81.05 &  & 67.88 &  & 84.50 & 98.06 & 99.38 &  & 66.38 & 89.34 & 94.60 & 64.13 \\ \midrule
\multicolumn{15}{c}{Large-scale Vision-Language pre-training} \\ \midrule
CLIP ViT-B/32 & 22.39 & 32.61 & 66.06 &  & 51.68 &  & 69.26 & 90.92 & 95.00 &  & 50.14 & 75.00 & 83.42 & 54.40 \\
CLIP ViT-B/16 & 23.67 & 33.96 & 68.68 &  & 53.84 &  & 71.56 & 91.40 & 95.82 &  & 52.30 & 76.78 & 84.64 & 56.46 \\
CLIP ViT-L/14 & 24.00 & 33.84 & 71.37 &  & 57.98 &  & 74.20 & 92.84 & 96.58 &  & 56.32 & 79.32 & 86.60 & 59.79 \\
VinVL (zero-shot) & 16.17 & 27.31 & 49.64 &  & 37.06 &  & 58.06 & 87.54 & 93.70 &  & 35.18 & 64.36 & 76.30 & 36.80 \\
VinVL & 34.56 & 44.39 & 83.35 &  & 75.74 &  & 88.14 & 98.32 & 99.40 &  & 74.66 & 92.58 & 96.34 & 75.73 \\
ViLT (zero-shot) & 23.22 & 33.64 & 65.34 &  & 58.24 &  & 77.10 & 95.12 & 97.92 &  & 56.70 & 82.54 & 89.58 & 57.26 \\
ViLT & 27.50 & 38.57 & 73.35 &  & 62.76 &  & 80.76 & 96.28 & 98.32 &  & 61.50 & 86.26 & 92.66 & 62.65 \\
BLIP & 36.02 & 44.56 & 88.50 &  & 82.68 &  & 92.28 & 98.96 & 99.58 &  & 81.90 & 95.38 & 97.80 & 82.32 \\ \midrule
\multicolumn{15}{c}{Different negative mining (NM) strategies} \\ \midrule
PVSE K=1, Sum triplet$^\dagger$ & 22.26 & 34.97 & 56.70 &  & 37.18 &  & 60.42 & 89.24 & 94.84 &  & 35.16 & 66.32 & 79.22 & 37.27 \\
PVSE K=1, SHM$^\dagger$ & 25.24 & 37.67 & 64.00 &  & 43.24 &  & 66.12 & 91.36 & 96.18 &  & 41.14 & 71.64 & 82.68 & 42.90 \\
PVSE K=1, HNM$^\dagger$ & 25.24 & 37.75 & 64.55 &  & 44.78 &  & 67.18 & 91.94 & 96.16 &  & 42.70 & 73.56 & 84.04 & 45.36 \\ \bottomrule
\end{tabular}
}
\vspace{.5em}
\caption{\small {\bf Re-evaluating VL models: Image-to-text retrieval results.}}
\label{tab:main_results_it}
\end{table}%

\begin{table}[t!]
\small
\centering
\setlength{\tabcolsep}{4pt}
\resizebox{\columnwidth}{!} {
\begin{tabular}{@{}lcccccccccccccc@{}}
\toprule
 & \multicolumn{3}{c}{ECCV} &  & CxC &  & \multicolumn{3}{c}{COCO 1K} &  & \multicolumn{4}{c}{COCO 5K} \\
 & mAP@R & R-P & R@1 &  & R@1 &  & R@1 & R@5 & R@10 &  & R@1 & R@5 & R@10 & PMRP \\  \midrule
\multicolumn{13}{c}{ResNet-152 \cite{resnet} image encoder + Bi-GRU \cite{cho2014properties} text encoder} \\ \midrule
VSE0$^\dagger$ & 30.41 & 40.48 & 66.37 &  & 21.56 &  & 39.95 & 74.71 & 84.71 &  & 19.78 & 46.12 & 59.87 & 42.36 \\
VSE++$^\dagger$ & 45.57 & 54.49 & 81.91 &  & 32.24 &  & 52.76 & 84.79 & 91.92 &  & 30.05 & 60.11 & 72.95 & 48.94 \\
PVSE K=1 & 44.55 & 53.41 & 83.93 &  & 32.88 &  & 53.49 & 85.11 & 92.08 &  & 30.64 & 61.37 & 73.62 & 48.32 \\
PVSE K=2 & 53.80 & 60.6 & 88.44 &  & 34.27 &  & 54.91 & 86.50 & 93.12 &  & 32.15 & 62.79 & 74.84 & 50.36 \\
PCME & 47.97 & 56.99 & 84.08 &  & 33.95 &  & 54.65 & 86.25 & 93.17 &  & 31.80 & 62.17 & 74.59 & 52.54 \\
PCME (CutMix)$^\dagger$ & 54.92 & 61.66 & 88.59 &  & 36.26 &  & 56.70 & 87.08 & 93.84 &  & 33.97 & 63.77 & 75.85 & 53.46 \\ \midrule
\multicolumn{13}{c}{Region features based on Bottom-up Attention \cite{Anderson2017up-down} and SCAN \cite{lee2018scan}} \\ \midrule
VSRN & 53.78 & 60.78 & 89.19 &  & 42.63 &  & 62.77 & 89.71 & 94.62 &  & 40.46 & 70.58 & 81.10 & 48.47 \\
VSRN + AOQ & 51.17 & 58.68 & 89.94 &  & 43.34 &  & 63.46 & 90.50 & 95.40 &  & 41.14 & 71.50 & 81.96 & 50.36 \\
CVSE & 46.59 & 54.88 & 84.16 &  & 38.71 &  & 59.88 & 89.29 & 94.81 &  & 36.60 & 68.04 & 79.57 & 50.74 \\
SGR & 44.35 & 52.93 & 86.49 &  & 42.39 &  & 62.06 & 89.56 & 94.90 &  & 40.47 & 69.64 & 80.28 & 51.84 \\
SAF & 44.55 & 53.07 & 85.66 &  & 42.18 &  & 62.24 & 89.53 & 94.98 &  & 40.12 & 69.73 & 80.38 & 52.42 \\
VSE infty (BUTD region) & 49.56 & 57.16 & 90.17 &  & 44.67 &  & 64.79 & 91.41 & 95.95 &  & 42.41 & 72.67 & 83.19 & 50.31 \\
VSE infty (BUTD grid) & 49.12 & 57.10 & 89.49 &  & 46.29 &  & 66.42 & 92.10 & 96.36 &  & 44.09 & 74.10 & 84.01 & 50.94 \\
VSE infty (WSL grid) & 50.02 & 57.45 & 91.82 &  & 53.70 &  & 72.04 & 93.92 & 97.19 &  & 51.64 & 79.34 & 87.58 & 51.16 \\ \midrule
\multicolumn{13}{c}{Large-scale Vision-Language pre-training} \\ \midrule
CLIP ViT-B/32 & 31.11 & 41.2 & 68.09 &  & 32.26 &  & 49.68 & 79.29 & 87.70 &  & 30.42 & 55.96 & 66.89 & 50.69 \\
CLIP ViT-B/16 & 34.82 & 44.01 & 73.42 &  & 34.67 &  & 52.47 & 80.86 & 88.90 &  & 33.07 & 58.42 & 68.99 & 51.80 \\
CLIP ViT-L/14 & 31.96 & 41.76 & 72.97 &  & 38.30 &  & 55.46 & 82.29 & 90.18 &  & 36.55 & 61.05 & 71.14 & 52.82 \\
VinVL (zero-shot) & 28.19 & 38.54 & 60.74 &  & 30.41 &  & 50.15 & 80.37 & 87.19 &  & 28.96 & 56.99 & 68.79 & 37.55 \\
VinVL & 47.06 & 54.71 & 92.19 &  & 59.77 &  & 76.62 & 95.15 & 97.91 &  & 58.12 & 83.20 & 90.08 & 46.79 \\
ViLT (zero-shot) & 30.46 & 39.97 & 72.65 &  & 42.45 &  & 62.26 & 90.70 & 96.21 &  & 40.56 & 70.00 & 81.07 & 51.90 \\
ViLT & 41.66 & 49.97 & 82.27 &  & 44.68 &  & 64.73 & 91.84 & 96.68 &  & 42.85 & 72.76 & 83.00 & 53.11 \\
BLIP & 45.01 & 52.30 & 93.47 &  & 65.92 &  & 79.95 & 95.80 & 97.82 &  & 64.32 & 85.73 & 91.57 & 53.15 \\ \midrule
\multicolumn{13}{c}{Different negative mining (NM) strategies} \\ \midrule
PVSE K=1, Sum triplet$^\dagger$ & 44.41 & 53.91 & 79.28 &  & 28.20 &  & 49.30 & 83.68 & 91.87 &  & 26.13 & 56.40 & 70.28 & 52.42 \\
PVSE K=1, SHM$^\dagger$ & 48.02 & 57.05 & 83.93 &  & 33.09 &  & 53.57 & 85.45 & 92.43 &  & 30.86 & 60.83 & 73.71 & 49.91 \\
PVSE K=1, HNM$^\dagger$ & 46.28 & 55.24 & 82.81 &  & 33.25 &  & 54.02 & 85.40 & 92.24 &  & 31.06 & 61.26 & 73.76 & 48.99 \\ \bottomrule
\end{tabular}
}
\vspace{.5em}
\caption{\small {\bf Re-evaluating VL models: Text-to-image retrieval results.}}%
\label{tab:main_results_ti}
\end{table}%

%% file: tables/model_performances_vs_machine_annotators.tex
\begin{table}[t]
\small
\centering
\setlength{\tabcolsep}{4pt}
\begin{subtable}[t]{\linewidth}
\centering
\begin{tabular}[t]{@{}lcccccc@{}}
\toprule
       & \multicolumn{6}{c}{Models used for annotations} \\
Models & PVSE  & VSRN & PCME & ViLT & CLIP & All  \\ \midrule
PVSE   & \textbf{76.5}  & 63.6 & 67.6 & 45.3 & 42.0 & 76.6 \\
VSRN   & 68.0  & \textbf{80.1} & 69.0 & 51.1 & 47.2 & \textbf{80.1} \\
PCME   & 67.7  & 64.3 & \textbf{77.3} & 46.0 & 44.1 & 77.4 \\
ViLT   & 59.5  & 59.8 & 58.8 & \textbf{62.0} & \textbf{49.3} & 72.4 \\
CLIP   & 51.7  & 51.5 & 52.6 & 42.8 & \textbf{49.3} & 64.3 \\
\bottomrule
\end{tabular}
\caption{\textbf{Text-to-Image Recall@1.}}
\label{tab:main_r1}
\end{subtable}
\begin{subtable}[t]{\linewidth}
\centering
\begin{tabular}[t]{@{}lcccccc@{}}
\toprule
       & \multicolumn{6}{c}{Models used for annotations} \\
Models & PVSE   & VSRN   & PCME  & ViLT  & CLIP  & All   \\ \midrule
PVSE   & \textbf{53.5}   & 39.4   & 42.1  & 30.1  & 31.3  & 42.1  \\
VSRN   & 41.8   & \textbf{55.9}   & 41.6  & 33.8  & 35.9  & \textbf{43.2}  \\
PCME   & 42.9   & 39.9   & \textbf{54.8}  & 31.8  & 33.8  & 43.1  \\
ViLT   & 34.0   & 34.5   & 33.9  & \textbf{45.5}  & 37.0  & 35.8  \\
CLIP   & 31.0   & 31.5   & 31.9  & 29.6  & \textbf{39.6}  & 32.1  \\ \bottomrule
\end{tabular}
\caption{\small {\bf Text-to-Image R-Precision.}}
\label{tab:main_pr}
\end{subtable}
\caption{\small {\bf Model performances vs. different annotation processes.} Each row indicates performances of the same model by different annotation strategies: using the annotations filtered by a specific model. For example, the first column of the tables shows the model performances by only using ``PVSE'' filtered annotation. ``All'' denotes the full annotations are used. The bold numbers denote the best model performances for each annotation strategy, where the best performed model and the model used for the annotation strategy are the same in all experiments.}
\label{tab:main}
\end{table}